\documentclass[journal]{IEEEtran}
\usepackage[T1]{fontenc}
\usepackage{graphicx}
\usepackage{times}
\usepackage{helvet}
\usepackage{courier}
\usepackage{amsmath}
\usepackage{algorithm}
\usepackage{algorithmic}
\usepackage{csquotes} 
\usepackage{color}
\usepackage{paralist}
\usepackage{amssymb}
\usepackage{indentfirst}
\usepackage{subfigure}
\usepackage{float}
\usepackage{multirow}
\usepackage{cite}
\usepackage{mathrsfs}

\usepackage{mathrsfs} 
\usepackage{amsfonts}
\usepackage{todonotes}
\usepackage{pgfplots} 
\usepackage{epstopdf}
\usepackage{epsfig}
\pgfplotsset{compat=newest}
\usepackage{color}
\definecolor{forestgreen}{RGB}{0,139,69}

\usepackage{xcolor}
\definecolor{citecolor}{HTML}{0071bc}
\usepackage[colorlinks, linkcolor=red,  anchorcolor=blue, citecolor=citecolor]{hyperref} 

\usepackage{xcolor}
\definecolor{SeaGreen4}{RGB}{0,205,102} 
\definecolor{SlateBlue}{RGB}{106,90,205} 
\definecolor{DarkRed}{RGB}{178,34,34} 
	
\usepackage[switch]{lineno}

\usepackage{textcomp,booktabs}
\usepackage{amssymb}
\usepackage{pifont}
\newcommand{\cmark}{\ding{51}}%

\usepackage{makecell}

\usepackage{colortbl}
\definecolor{mygray}{gray}{.9}
\definecolor{mypink}{rgb}{.99,.91,.95}
\definecolor{mycyan}{cmyk}{.3,0,0,0}

\begin{document}

\title{Pedestrian Attribute Recognition via CLIP based Prompt Vision-Language Fusion }

\author{Xiao Wang, \emph{Member, IEEE}, Jiandong Jin, Chenglong Li*, Jin Tang, Cheng Zhang, Wei Wang    
\thanks{Xiao Wang, Jin Tang, and Cheng Zhang are with the School of Computer Science and Technology, Anhui University, Institute of Artificial Intelligence, Hefei Comprehensive National Science Center, Hefei 230601, China. (email: xiaowang@ahu.edu.cn, tangjin@ahu.edu.cn, cheng.zhang@ahu.edu.cn)} 
\thanks{Jiandong Jin, Chenglong Li are with Information Materials and Intelligent Sensing Laboratory of Anhui Province, Anhui Provincial Key Laboratory of Multimodal Cognitive Computation, the School of Artificial Intelligence, Anhui University, Hefei 230601, China. (email: jdjinahu@foxmail.com, lcl1314@foxmail.com)}
\thanks{Wei Wang is with Video Investigation Detachment of Hefei Public Security Bureau, Hefei 230031, China. (email: yzyzww@sohu.com)}
\thanks{* Corresponding Author: Chenglong Li} 
}

\markboth{ IEEE Transactions on Circuits and Systems for Video Technology } 
{Shell \MakeLowercase{\textit{et al.}}: Bare Demo of IEEEtran.cls for IEEE Journals}

\maketitle

\begin{abstract}
Existing pedestrian attribute recognition (PAR) algorithms adopt pre-trained CNN (e.g., ResNet) as their backbone network for visual feature learning, which might obtain sub-optimal results due to the insufficient employment of the relations between pedestrian images and attribute labels. 
In this paper, we formulate PAR as a vision-language fusion problem and fully exploit the relations between pedestrian images and attribute labels. Specifically, the attribute phrases are first expanded into sentences, and then the pre-trained vision-language model CLIP is adopted as our backbone for feature embedding of visual images and attribute descriptions. The contrastive learning objective connects the vision and language modalities well in the CLIP-based feature space, and the Transformer layers used in CLIP can capture the long-range relations between pixels. Then, a multi-modal Transformer is adopted to fuse the dual features effectively and feed-forward network is used to predict attributes. 
To optimize our network efficiently, we propose the region-aware prompt tuning technique to adjust very few parameters (i.e., only the prompt vectors and classification heads) and fix both the pre-trained VL model and multi-modal Transformer. 
Our proposed PAR algorithm only adjusts 0.75\% learnable parameters compared with the fine-tuning strategy. It also achieves new state-of-the-art performance on both standard and zero-shot settings for PAR, including RAPv1, RAPv2, WIDER, PA100K, and PETA-ZS, RAP-ZS datasets. The source code and pre-trained models will be released on \url{https://github.com/Event-AHU/OpenPAR}. 
\end{abstract}

\begin{IEEEkeywords}
Pedestrian Attribute Recognition, Pre-trained Big Models, Prompt Learning, Multi-Modal Fusion, Vision-Language 
\end{IEEEkeywords}

\IEEEpeerreviewmaketitle

\section{Introduction}

\IEEEPARstart{P}{edestrian} attribute recognition (PAR)~\cite{wang2022PARsurvey} aims to describe the middle-level semantic information of a person using a set of pre-defined attributes, such as~\emph{age, height, hairstyle, clothing}. It plays an important role in the computer vision community, especially intelligent video surveillance and automatic drive, and also facilitates the research of other visual tasks, including person re-identification~\cite{chai2022videoREIDAttribute}, person search~\cite{wang2022align}, and pedestrian detection~\cite{tian2015pedestriandetection}. With the help of artificial intelligence, such as the CNN (Convolutional Neural Network), and RNN (Recurrent Neural Network), this research field has received much attention and achieved great development~\cite{he2016resnet, tealab2018time}. However, it is still challenging due to poor imaging quality in extreme scenarios, including motion blur, shadow, occlusion, low resolution, multiple views, and nighttime.

\begin{figure*} 
\centering
\small
\includegraphics[width=0.9\textwidth]{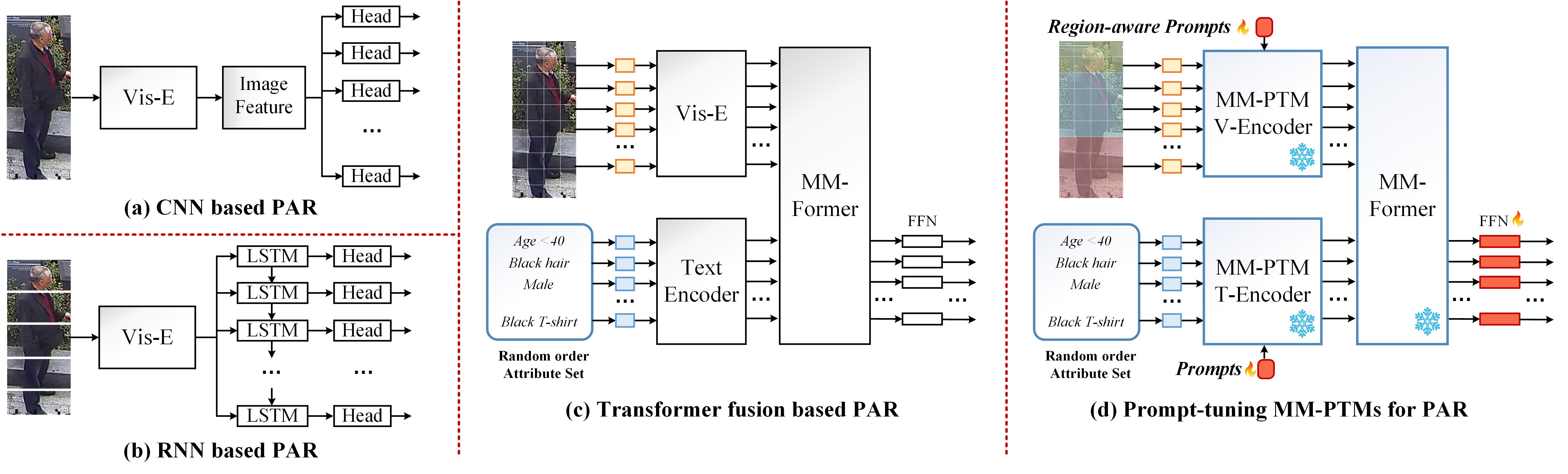}
\caption{Comparison of CNN-based, RNN-based, Transformer-based, and our newly proposed CLIP guided vision-language fusion frameworks for pedestrian attribute recognition.} 
\label{frontImg} 
\end{figure*}

Current PAR algorithms focus on CNN-based models~\cite{zhang2014panda, abdulnabi2015multitask} and RNN-based models~\cite{wang2017JRL, zhao2018GRL}, whose frameworks are shown in Fig.~\ref{frontImg} (a) and (b) respectively. CNN-based PAR models attempt to learn a mapping from pedestrian image to pre-defined attributes. Some researchers consider the semantic relations between pedestrian attributes and introduce RNN to sequentially infer the attributes. Although these works achieve good performance on existing benchmark datasets, these models still fail to achieve efficient and high-performance recognition. We think these issues might be caused by the following issues. 
1) The semantic information of attribute phrases is rarely exploited but is very important for attribute recognition. 
2) The relations between semantic attributes and visual representations are insufficiently explored and exploited. 
3) Existing works usually adopt CNN for feature learning which encodes the local relations only, but some attributes heavily rely on long-range pixel relationship modeling, e.g., \textit{body shape} and \textit{dress style}.

Recently, the success of self-attention and Transformer~\cite{vaswani2017Former} networks in natural language processing and computer vision has attracted more and more attention. Some studies reveal that the Transformer captures the long-range relations between input tokens well. Many works are developed by using Transformer independently or concatenating with CNN models, such as TransT~\cite{chen2021TransT} in visual tracking and DETR~\cite{carion2020DETR} in object detection. 

Cheng et al. introduce the Transformer into the PAR task and propose a simple baseline VTB~\cite{cheng2022VTB} by fusing visual-text inputs, and the structure is shown in Fig.~\ref{frontImg} (c). More in detail, they extract the deep visual features using ViT backbone network~\cite{dosovitskiy2020VIT}, and then split and expand the attribute phrases into sentences. Then, a visual-text Transformer is introduced to fuse the dual modalities for attribute recognition. The VTB addresses the aforementioned issues to some extent and achieves better results than previous works, but it still ignores the following key clues for PAR. 
\textbf{First}, it simply introduces pre-trained ViT~\cite{dosovitskiy2020VIT} and BERT~\cite{kenton2019bert} for visual and text feature extraction independently. Therefore, the mining of internal relations might be weak. 
\textbf{Second}, it designs a multi-modal Transformer to fuse the visual-text features and adopts fine-tuning for the optimization of the whole network. The huge amount of learnable parameters will put forward higher requirements for GPU computing power. 
The aforementioned two questions inspire us to think about how to carry out PAR with high accuracy and efficiency.

To relieve the power of Multi-Modal Pre-Trained big Models (MM-PTM) in PAR task, in this paper, we formulate PAR as a multi-modal fusion problem and propose a novel CLIP~\cite{radford2021CLIP} based PAR framework to make full use of prior knowledge obtained from the pre-trained CLIP model. As shown in Fig.~\ref{PromptPARframework}, we first expand the attribute phrases into language descriptions using a prompt template like ``\emph{a pedestrian whose \underline{~~~~~~} is \underline{~~~~~~}}''. For example, \emph{\textbf{long hair}} $\longrightarrow$ \emph{a pedestrian whose \underline{\textbf{hair}} is \underline{\textbf{long}}}.  
Then, we feed both pedestrian image and text into a pre-trained vision-language model (CLIP~\cite{radford2021CLIP} adopted in our experiments) for feature embedding. Next, a multi-modal Transformer is introduced to fuse the dual features and a couple of feed-forward layers are used for attribute prediction. Unlike previous works that adopt fine-tuning for network optimization, we propose an effective region-aware prompt-tuning technique to achieve efficient optimization of the whole network.  
 Specifically, we propose different prompt tokens for various body parts with the assumption that these local prompts help the spatial-aware attribute feature learning. As human attributes like the \textit{black hair} and \textit{glasses} are more relevant to the head region, while \textit{T-shirt} is more relevant to the upper body region. Thus, it may be helpful if we inject different prompts instead of only one global prompt used in other works~\cite{2022vpt}. The parameters of both pre-trained vision-language models and multi-modal Transformers are all fixed, i.e., only the prompts and recognition heads are learnable. Region-aware visual prompts and text prompts are randomly initialized and then concatenated with the visual and language features separately. Our experiments demonstrate such prompt tuning improves training efficiency significantly while maintaining the overall recognition performance. 


To sum up, the contributions of this paper can be concluded as the following three aspects. 

\begin{itemize}
\item We propose a novel CLIP-guided vision-language fusion framework for pedestrian attribute recognition, termed PromptPAR, which takes full advantage of powerful feature representations of pre-trained vision-language models to connect the relations between pedestrian images and attribute labels. Moreover, it handles the imbalanced data distribution of attributes and obtains a more generalized PAR model. To our best knowledge, it is the first work to exploit the pre-trained vision-language big models for the PAR task.

\item We propose an effective region-aware prompt tuning strategy to achieve more efficient training of our proposed PAR network. It fixes the parameters of both the pre-trained big model and the vision-language fusion module and thus adjusts 0.75\% learnable parameters compared with the fine-tuning strategy.

\item Extensive experiments on multiple PAR benchmark datasets (i.e., RAPv1~\cite{2016rapv1}, RAPv2~\cite{2019rapv2}, PETA~\cite{deng2014peta}, PA100K~\cite{2017pa100k}, WIDER~\cite{li2016wider}) validate the effectiveness of our PromptPAR. Specifically, PromptPAR achieves 90.15\%, 82.38\%, 81.00\%, and 92.0\% on the F1 score of PA100K, RAPv1, RAPv2, and mA of the WIDER dataset, respectively, which are all new state-of-the-art performances. Our PromptPAR also sets new SOTA results on two zero-shot PAR datasets and improves the baseline up to +5.52\% and +5.66\% on the accuracy metric on PETA-ZS and RAP-ZS datasets, respectively.
\end{itemize}

\section{Related Work} 

In this section, we give an introduction to pedestrian attribute recognition, prompt learning, and multi-modal pre-trained models. More related works on PAR can be found in the following survey~\cite{wang2022PARsurvey} and paper list~\footnote{\url{github.com/wangxiao5791509/Pedestrian-Attribute-Recognition-Paper-List}}.

\noindent 
\textbf{Pedestrian Attribute Recognition.~} 
Current PAR algorithms can be divided into the following categories, including CNN-based, RNN-based, attention-based, and curriculum learning-based methods~\cite{zhang2014panda, zhao2018GRL, jia2022learning, dong2017multicurriculum, guo2022visual}. Specifically, CNN-based PAR models use CNN to extract deep feature representations of pedestrian images and adopt fully connected layers as the classifiers for final recognition. Zhang et al. propose the PANDA~\cite{zhang2014panda} which combines the part-based model and human attribute classification based on CNN. A multi-task CNN is proposed in the work~\cite{abdulnabi2015multitask} to predict the human attributes in a binary classification manner, which makes the CNN shares the visual information between different attributes. RNN-based PAR algorithms consider the semantic correlations between human attributes and attempt to predict attributes in a more intelligent way. For example, Wang et al. propose an end-to-end RNN model to learn sequential semantic correlations between pedestrian attributes~\cite{wang2017JRL}. Zhao et al. propose the GRL~\cite{zhao2018GRL} to exploit the potential dependencies between pedestrian attributes by considering intra-group attribute mutual exclusion and inter-group attribute association. 
With the popularity of deep learning in the PAR community, the attention mechanism is also considered for PAR. To be specific, Guo et al. use two visual attentional consistency modules to increase the credibility of attention maps~\cite{guo2022visual}. Jia et al.~\cite{jia2022learning} propose the One-specific-Feature-for-One-Attribute (OFOA) mechanism for PAR which is a distinguishable and highly robust pedestrian attribute feature learning framework. Inspired by the human learning mechanism, some researchers propose curriculum learning based PAR algorithms that can realize smarter recognition in an easy-to-hard way. MTCT~\cite{dong2017multicurriculum} is a multi-task course transfer network that can achieve this goal without manual labeling. 
Different from these works, our proposed PromptPAR is developed based on a vision-language fusion framework. It is firstly proposed in VTB~\cite{cheng2022VTB}, however, this work adopts separately trained visual and text backbones, and fuses dual modalities using finetuning. Our framework adopts the pre-trained VL models for input embedding which bridges the dual modalities more effectively. Also, we adopt prompt tuning which improves the baseline in both the training efficiency and recognition performance.

\noindent 
\textbf{Prompt Learning.~} 
To better utilize the pre-trained big models, prompt learning is proposed to transform the downstream tasks to have similar targets. It is a novel tuning method that is significantly different from the widely used fine-tuning, therefore, prompt learning is also termed the fourth paradigm in the natural language processing field. Many works are developed based on this idea, such as the GPT-3~\cite{2020gpt3}. More importantly, prompt learning is also widely exploited in the multi-modal community. Wang et al.~\cite{Wang2021actionclip} formulate the action recognition as a multi-modal video-text matching problem and propose the ActionCLIP. VPT~\cite{2022vpt} is proposed to decrease the number of parameters that need to be fine-tuned by injecting a set of learnable prompts into the input tokens. Zhou et al.~\cite{zhou2022coop} propose a simple method termed Context Optimization (CoOp) which can adapt the VL models for image recognition efficiently. They also introduce the CoCoOp~\cite{zhou2022cocoop}, a method for generating input-conditional tokens for each image, aimed at resolving the issue of limited generalization when applying learned context to unseen labels. Gao et al.~\cite{gao2023clipadapter} employed Adapters to learn new features and blended them residually with the original features. This approach allows for the simultaneous optimization of the textual and visual branches. Guo et al.~\cite{guo2023calip} enabled the two modalities, visual and language, to interact during encoding, resulting in a versatile visual model without the need for additional parameters. 
Zhang et al.~\cite{zhang2021tip} introduced Tip-Adapters, a training-free method that demonstrates strong few-shot classification performance by directly configuring the Adapter weights using a caching model. 

Inspired by these works, we propose the visual-text prompt tuning to adapt the pedestrian attribute recognition task to the CLIP model. It reduces the number of parameters needed to be trained and also brings us a better recognition performance compared with the standard fine-tuning algorithm.

\noindent 
\textbf{Multi-modal Pre-trained Models.~} 
Current multi-modal big models are usually pre-trained on vision-language data, since these data are relatively easy to collect and can be used in a wide range of scenarios~\cite{wang2022MMPTMSurvey}. A large number of pre-trained multi-modal models are developed based on Transformers architecture~\cite{vaswani2017Former}. 
To be specific, Li et al.~\cite{2019VisualBERT} propose a single-stream pre-trained model by aligning the regions of the input text and images through the Transformer's self-attention mechanism. Li et al. propose the Oscar~\cite{li2020oscar} which aligns the image and language by using object labels detected in the image as anchor points and achieves a better performance. CLIP~\cite{radford2021CLIP} is obtained by conducting contrastive learning between dual modalities on 400 million image-text pairs. 
Additionally, extensive research has been conducted on multi-modal training models in the 3D multi-modal domain. For instance, PointCLIP~\cite{zhang2022pointclip}, PointCLIPv2~\cite{zhu2023pointclip2}, ULIP~\cite{xue2023ulip}, and ULIP2~\cite{xue2024ulip2} align point clouds, images, and text into a unified feature space by cross-modal matching. Joint-MAE~\cite{guo2023joint} establishes an implicit semantic and geometric correlation between 2D and 3D by using a joint masking mechanism. Binding models such as ImageBind~\cite{girdhar2023imagebind}, Point-Bind~\cite{guo2023point}, and LanguageBind~\cite{zhulanguagebind} suggest that humans bind different modalities through specific channels. However, collecting complete multimodal data is challenging. These approaches achieve multi-modal semantic alignment by learning a joint embedding space using paired data from one modality and any other modality.
Inspired by the success of multi-modal pre-trained models in many downstream tasks, in this work, we propose the first multi-modal pre-trained models based pedestrian attribute recognition framework. Together with prompt tuning, our proposed PromptPAR shows its advantages in both training efficiency and overall recognition performance.

\section{Methodology}  
In this section, we first give a review of the pre-trained model CLIP~\cite{radford2021CLIP} and an overview of our proposed PromptPAR framework. Then, we talk about the detailed network architecture, including the visual feature and attribute feature extraction, visual-text fusion Transformer, and prompt tuning module. After that, we will discuss the loss functions used for the optimization of our framework.

\subsection{Preliminary and Overview} 

\begin{figure}  
\centering
\small
\includegraphics[width=0.48\textwidth]{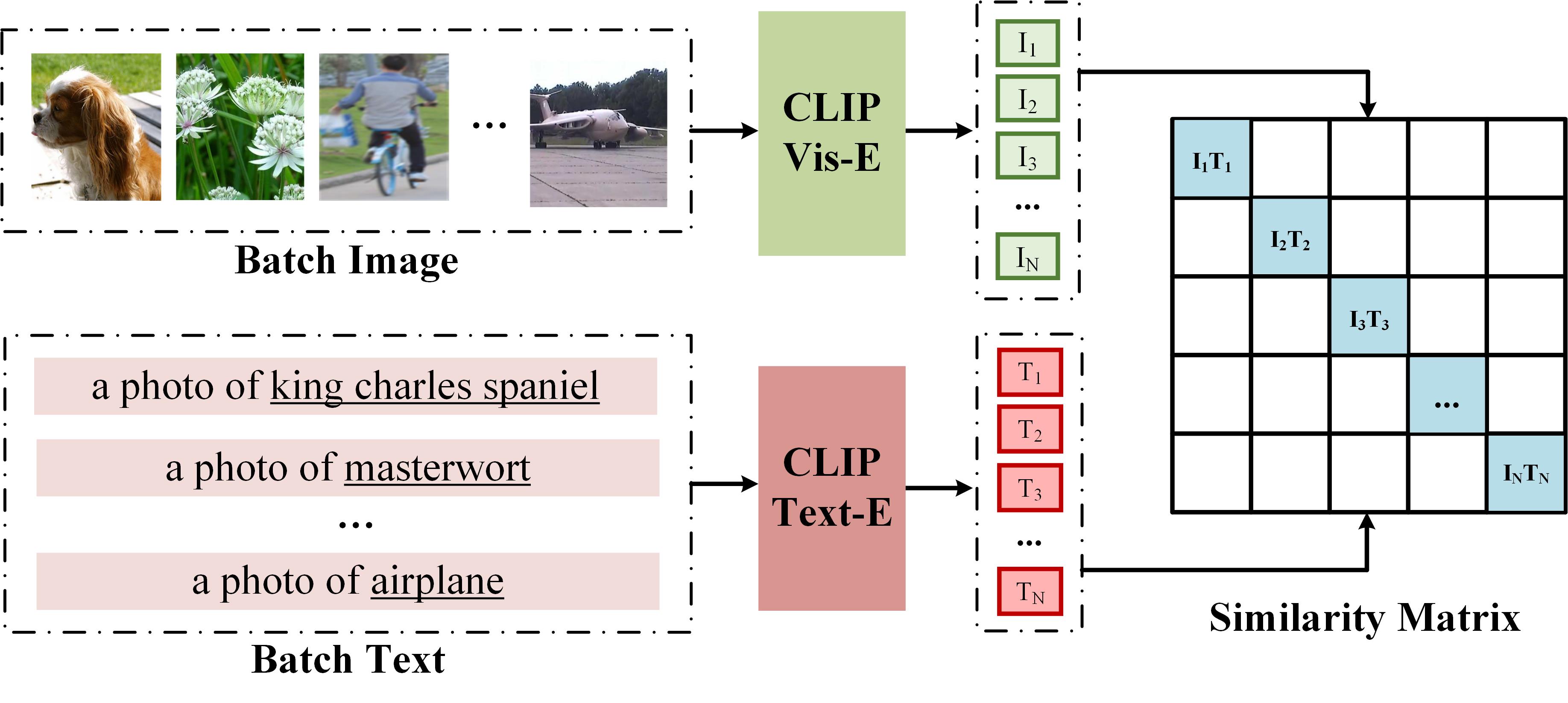}
\caption{An illustration of CLIP model~\cite{radford2021CLIP}. It takes a batch of image-text pairs as input and encodes their features using ResNet/Transformer network. The contrastive learning loss is built upon the vision-language paired or unpaired samples. It shows a great zero-shot transfer learning ability on many downstream tasks. 
}  
\label{CLIPmodel}
\end{figure}

\textbf{Preliminary of CLIP Model.~}  
Considering the high-quality feature representation and zero-shot transfer learning ability, in this paper, we adopt the pre-trained VL models for attribute recognition. Specifically, the CLIP is selected in our experiments due to its simplicity and good performance as already validated in many tasks.
This model is pre-trained on a dataset consisting of 400 million image-text pairs. As shown in Fig.~\ref{CLIPmodel}, the CLIP contains two branches, i.e., the visual and text encoders. The Transformer network is adopted as the encoder of both modalities. In addition, they also use the ResNet~\cite{he2016resnet} as the visual encoder to get multiple versions of pre-trained models. Once the two features are extracted, the cosine similarity between them is calculated. The target of the learning is to pull the matched visual-text pairs closer and push the unmatched pairs farther. It is worth noting that the CLIP also follows the \emph{prompt engineering} to provide diverse prompts to address the issues caused by ambiguous language descriptions. In the inference stage, the semantic distance between visual and text samples can be easily obtained by feeding them into the CLIP encoders. In our experiments, the CLIP ViT-B/16 and CLIP ViT-L/14 are all adopted for evaluation. More details can be found in our experiments in Section~\ref{experiments}.

\begin{figure*}[!htp] 
\centering
\small
\includegraphics[width=0.85\textwidth]{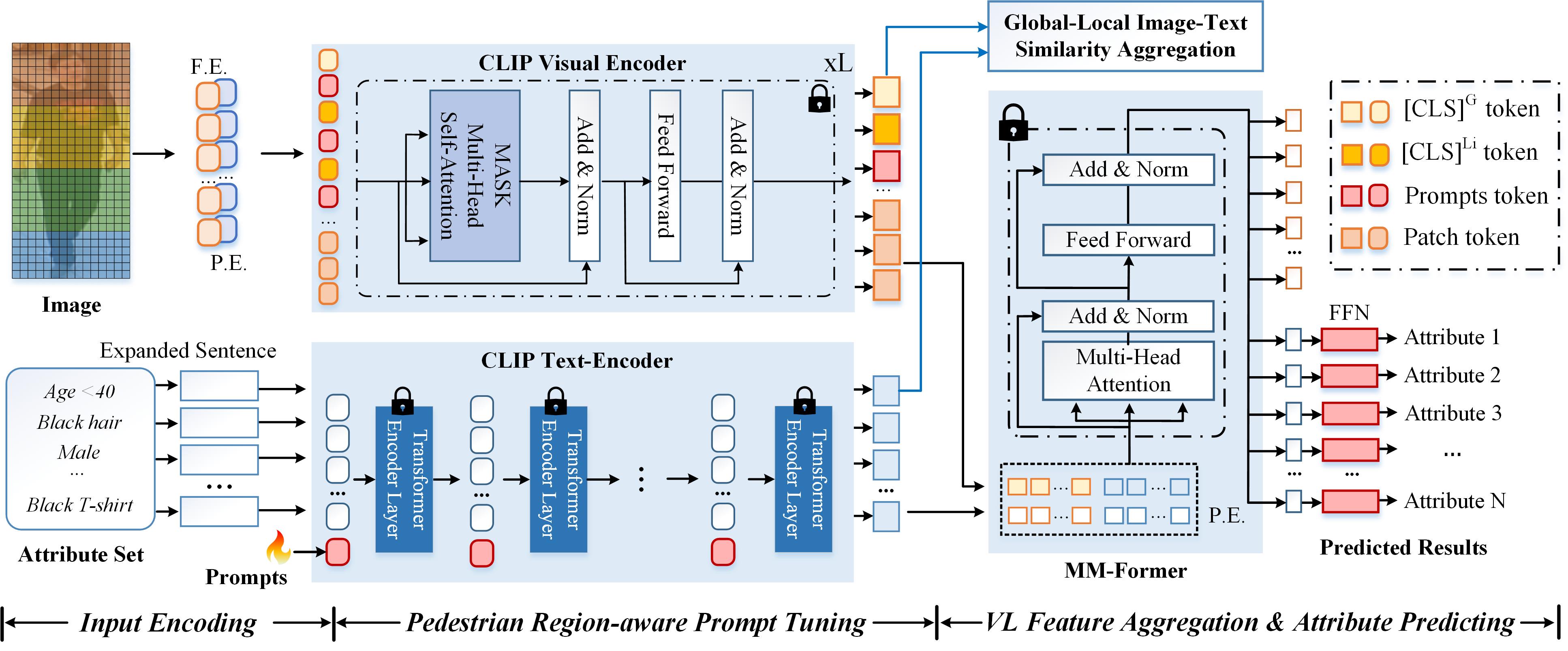}
\caption{\textbf{An illustration of our proposed PromptPAR framework which takes the pedestrian image and pre-defined attribute set as input and models the PAR task as a vision-language fusion problem.} It contains three main modules, including the CLIP visual encoder, CLIP textual encoder, multi-modal Transformer (MM-Former), and classification head. The utilization of CLIP encoders brings us a better feature representation and the MM-Former outputs a unified feature representation for attribute classification. More importantly, we adopt prompt tuning to optimize very few network parameters, in other words, only the prompt vectors and classification head are tunable. Extensive experiments demonstrate the efficiency and effectiveness of our proposed PAR framework. 
}
\label{PromptPARframework}
\end{figure*}

\textbf{Overview of Our Proposed PromptPAR.~} 
In this paper, we treat the PAR as a multi-modal fusion problem, as shown in Fig.~\ref{PromptPARframework}, our proposed PromptPAR framework consists of four main modules, i.e., the visual encoder, textual encoder, multi-modal Transformer (MM-Former), and classification head. To better exploit the vision-language relations, we resort to the pre-trained CLIP for the input embedding. Specifically, the CLIP ViT-L/14 is used for visual feature encoding, which takes equal-sized patches from the pedestrian image as the input. For the text input, we first segment the given attributes into separate words, then, expand each attribute into a sentence using the prompt template. The text encoder takes the sentence as input and output language feature representation. The visual and text features are concatenated together and fed into the multi-modal Transformer. Finally, the classification head is introduced to project the obtained features into corresponding attribute response scores. Different from the regular fine-tuning for the optimization of neural networks, in this work, we propose to optimize the network parameters as little as possible with prompt tuning to \emph{prevent over-fitting} and \emph{maintain the vision-language feature space of CLIP as best as possible}. Extensive experiments on multiple PAR benchmark datasets validate that prompt tuning improves training efficiency and overall accuracy significantly. More implementation details can be found in the following sub-sections.

\subsection{Network Architecture} 

Given a pedestrian image $\mathcal{I} \in \mathbb{R}^{H\times W \times C}$, the goal of pedestrian attribute recognition is to predict a set of attributes the person has from a pre-defined attribute list $\mathcal{A} = \{A_1, A_2, ..., A_N\}$. The $H, W$, and $C$ are the height, width, and channel of image $I$, $N$ is the number of 
all the defined attributes. In this work, we formulate the attribute prediction as a CLIP-guided vision-language fusion problem and optimize our framework using prompt tuning.

\textbf{Visual Branch. } 
Existing pedestrian attribute recognition models usually adopt CNN as their backbone network and are trained on ImageNet~\cite{deng2009imagenet} dataset. Obviously, their visual features are easily limited by the following issues:
1). The receptive field is limited by local convolutional filters and only local relations between pixels are modeled well using CNN. 
2). The weights pre-trained using the classification task on the ImageNet dataset may have insufficient ability to learn the features and relationships of visual-language data.

In this paper, the visual encoder of pre-trained CLIP~\cite{radford2021CLIP} is adopted as our backbone, which alleviates the aforementioned two issues effectively. The self-attention mechanism in CLIP visual encoder captures the long-range relations between pixels well and the pre-training on the image-language dataset builds the connections between the dual modalities. 
In our implementation, the input pedestrian image $\mathcal{I} \in \mathbb{R}^{H\times W \times C}$ is divided into $m$ patches with a fixed size, therefore, a set of patches $\mathcal{X}=\{x_1, x_2, ..., x_m\}$ can be obtained, where $x_j \in \mathbb{R}^{P \times P \times C}$, $P$ is the scale of each patch. Then, an embedding matrix $\mathcal{E}$ is used to project each patch into token representations and added with  pre-trained CLIP position embedding $PE \in \mathbb{R}^{P \times P \times d}$ to get the input tokens of the visual encoder, i.e., $\mathcal{F}_0^V = \mathcal{X*E} + PE$. $d$ is the dimension of input tokens of the visual encoder, note that the positional encoding here stays frozen during training. The input tokens are concatenated with a learnable $d$ dimension classification token $\mathbf{CLS}$ and fed into the Transformer layers, specifically,
\begin{equation}
\label{clip_vis_vit} 
[\mathbf{CLS}_i,  \mathcal{F}_{i}^V] = L_i([\mathbf{CLS}_{i-1},  \mathcal{F}_{i-1}^V]) 
\end{equation}  
where $[\cdot, \cdot]$ denotes the concatenate operation. $L_i$ denotes the $i$-th layer of the Transformer encoder.

In this procedure, we freeze the parameters of both visual and language encoders of CLIP to make the training procedure more efficient and hardware-friendly. That is to say, no changes are made to the already learned vision-language feature space of CLIP model. In this work, multi-modal prompt tuning is proposed to train our framework more efficiently. The visual prompt vector $\mathcal{P}^V$ is generated as part of the input. More details about the prompt tuning will be described in Section \ref{prompttuning}.

\textbf{Textual Branch. } 
In this work, we formulate the PAR as a vision-language fusion problem to better use the pre-trained CLIP model. For the pedestrian attributes $\mathcal{A} = \{A_1, A_2, ..., A_N\}$, we split and expand each of them into language descriptions by following VTB~\cite{cheng2022VTB}. Let's take the attribute \emph{`age31-45'} as an example, we first split this attribute into phrase formulation, i.e., \emph{`age 31 to 45'}. Then, the prompt engineer is adopted to expand the phrase into language description $\mathcal{S} = \{W_1, W_2, ..., W_N\}$, each $W$ denotes one English word, i.e., \emph{age 31 to 45} $\longrightarrow$ \emph{A pedestrian whose \underline{age} is \underline{31 to 45}}. Note that, all the pedestrian attributes are needed to be processed in such a split-expand way.

Once we obtained all the language descriptions of pedestrian attributes, the ViT-L/14 of CLIP~\cite{radford2021CLIP} is adopted as the text encoder for textual feature embedding and learning. It contains 12 Transformer blocks and transforms the language description of each attribute $\mathcal{S}$ into a token whose dimension is 768. 
Then, the textual prompt is implanted into the input features for joint visual-language prompt tuning, as shown in Fig.~\ref{PromptPARframework}. By feeding them into the CLIP text encoder, we can finally get the textual features $\mathcal{F}^T$.

\textbf{MM-Former. } 
The aforementioned feature embedding procedure of pedestrian image and attributes can be seen as the underlying modeling of dual modalities. In this work, we also propose to model their relations in an explicit way. As shown in Fig.~\ref{PromptPARframework}, a multi-modal Transformer (short for MM-Former) is proposed to take the dual modalities as input. To be specific, the text features are projected into the same dimension as visual features using a linear layer $\phi(\cdot)$. Then, the two features are concatenated into a unified feature representation 
$\mathcal{F} = [\phi(\mathcal{F}^T), \mathcal{F}^V]$.

Note that the MM-Former only contains one Transformer layer and its parameters are initialized using the last layer of the ViT-B/16 model which is pre-trained on ImageNet-21K~\cite{deng2009imagenet} and fine-tuned on ImageNet-1K~\cite{deng2009imagenet}. Following VTB~\cite{cheng2022VTB}, in this paper, we also select the textual features $Z$ obtained from the MM-Former network for attribute recognition to enhance the text semantics and eliminate the effect of attribute order on the prediction results. A Feed Forward Network ($FFN$) is adopted as the classification head for attribute score regression $\mathcal{R}$:
\begin{equation}\label{predict}
\mathcal{R}=FFN(Z)=\sigma(\textbf{w}Z+b)
\end{equation}
where $\textbf{w} \in \mathbb{R}^{N \times d_{out}}$ and $b$ is the weight and bias of the $FFN$. $\sigma(\cdot)$ is a sigmoid function.

\subsection{Prompt Tuning} \label{prompttuning}
Based on the aforementioned pre-trained CLIP model and MM-Former, we can directly train the whole network using standard finetuning. However, it will put forward a great demand for computing power resources. Inspired by the recent downstream tasks based on the pre-trained big models, such as the VPT~\cite{2022vpt} which fix most of their parameters and only tune very few ones for efficient training, in this work, we propose a novel vision-language prompt tuning algorithm to optimize our framework. To be specific, a set of learnable prompt tokens is generated and concatenated with input features. Similar to the VPT model~\cite{2022vpt}, we also tried the shallow and deep versions of prompt tuning for the vision and language branch, termed VLP-shallow and VLP-deep, respectively.

The VLP-shallow embeds a set of learnable prompt tokens into the first Transformer layer only, whose dimension is the same as the input token. Formally, we denote the visual and language prompt tokens as $\mathcal{P}^V =  \{ p^V_k \in \mathbb{R}^{1024}$\}, 
$\mathcal{P}^T = \{ p^T_k \in \mathbb{R}^{768}$\}, the $p^V_k$ and $p^T_k$ denote each prompt token. 
Therefore, we have: 
\begin{align}\label{promptdeep}
&[\mathbf{CLS}_1,\mathcal{Z}_1,\mathcal{F}_{1}^V]=L_1([\mathbf{CLS}_0,  \mathcal{P}^V,  \mathcal{F}_{0}^V]) \\
&[\mathcal{F}_{1}^T,\mathcal{Z}_1]=L_1([ \mathcal{F}_{0}^T,\mathcal{P}^T])
\end{align}
For the VLP-deep prompt, a set of learnable prompt tokens is generated and integrated into multiple Transformer layers. In addition to the vision and language prompt tokens, we also tune the parameters of FFN. More in detail, the prompt token embedding in the visual/textual feature extractor can be expressed as: 
\begin{equation}\label{promptDeepVIS}
[\mathbf{CLS}_i, \_, \mathcal{F}_{i}^V] = L_i([\mathbf{CLS}_{i-1}, \mathcal{P}_{i-1}^V, \mathcal{F}_{i-1}^V])
\end{equation}
\begin{equation}\label{promptDeepText}
[\mathcal{F}_{i}^T, \_] = L_i([\mathcal{F}_{i-1}^T, \mathcal{P}_{i-1}^T])
\end{equation}
Once we integrate the prompt features into the visual and text features, we concatenate and feed the two features into the MM-Former network. Thanks to the vision-language prompt tuning scheme used in this work, our model achieves new state-of-the-art performance on multiple PAR benchmark datasets. Also, we achieve a higher training efficiency when optimizing partial of our model only.

\subsection{Region-aware Prompt Tuning} \label{PRPT} 
In addition to the aforementioned global prompts, in this work, we also exploit the region-aware prompt tuning (RegionPTune) strategy to learn different prompts for various pedestrian regions. The motivation of our region-aware prompts can be summarized as two points: 
1). Transformer focuses on long-range global representation learning which will make the injected multiple prompts share similar activation responses. Our visualization of attention maps between the prompt tokens and human image tokens fully validated our assumption, as shown in Fig.~\ref{respond}. We find that the global prompt will pay too much attention to the partial human regions. 
2). The detected pedestrian images can be roughly divided into four main parts, i.e., the head, the upper body, the lower body, and the feet region. And many human attributes are corresponding to these regions well. 
These observations and reflections inspired us to design region-aware prompts for better pedestrian feature learning.

To be specific, we divide the patches into $K$ regions from top to bottom, each of which is then assigned Region-aware Prompts. 
We divide the prompts as $P^V = \{P^G, P^{L_1}, ... , P^{L_K}\}$, where $P^G=\{ p_1^G , p_2^G ,..., p_m^G \}$ represents the prompt responsible for learning the complete pedestrian picture information, while $P^{L_j}=\{ p_1^{L_j}, p_2^{L_j}, ... , p_m^{L_j} \}$ corresponds to the prompt that interacts exclusively with the \textit{j}-th pedestrian region. Here, $m$ denotes the number of prompts for each region. Therefore, we can re-formulate the Eq.~\ref{promptDeepVIS} as follows: 
\begin{equation}
\begin{aligned}
&[\mathbf{CLS}_i^G, \_, \ldots, \mathbf{CLS}_i^{L_j}, \_, \ldots, \mathcal{F}_{i}^V] \\
&= L_i([\mathbf{CLS}_{i-1}^G, \mathcal{P}_{i-1}^G, \ldots, \mathbf{CLS}_{i-1}^{L_j}, \mathcal{P}_{i-1}^{L_j}, \ldots, \mathcal{F}_{i-1}^V])
\end{aligned}
\end{equation} 

Note that, we adopt the widely used masking strategy for local prompts to make the Transformer only focus on one region of human image. We follow the Global-Local Image-Text Similarity Aggregator~\cite{cdul2023} to adjust the prompts to generate the visual region features that match the attribute representations well. The similarity between region CLS tokens and the attribute representations can be calculated and used to denote the confidence of corresponding attributes. In addition, we can get such similarity matrices between global CLS tokens and attributes. We average these two similarity matrices and adopt the ground truth to guide such distance metric learning via cross-entropy loss function GLloss $\mathcal{L}_{GL}$, i.e., 
\begin{equation}\label{loss_gl}
\mathcal{L}_{GL}=-\frac{1}{M}\sum_{i=1}^M\sum_{j=1}^N(y_{ij}\log(p_{ij}^{GL})+(1-y_{ij})\log(1-p_{ij}^{GL}))
\end{equation}

Given the pedestrian attributes predicted by our model and the ground truth labels, we adopt the weighted cross-entropy loss function~\cite{deepmar} to measure the quality of convergence. This loss function considers the distributions of each attribute and can handle the imbalanced data to some extent. Formally, we have: 
\begin{equation}\label{loss_cls}
\mathcal{L}_{CLS}=-\frac{1}{M}\sum_{i=1}^M\sum_{j=1}^Nw_j(y_{ij}\log(p_{ij})+(1-y_{ij})\log(1-p_{ij}))
\end{equation}
where $w_j=e^{-r_j}$ is the imbalance weight of the $j$-th attribute, $r_j$ is the proportion of occurrences of the $j$-th attribute in the training subset. $p_{ij}$ and $y_{ij}$ denotes predicted attributes and ground truth, respectively. 
The final overall loss is computed as follows:
\begin{equation}\label{loss_all}
\mathcal{L} = \mathcal{L}_{CLS} + \alpha \mathcal{L}_{GL}
\end{equation}
where $\alpha$ is a tradeoff parameter between the two loss functions and we experimentally set it as 0.5 in our experiments.

\section{Experiments}  \label{experiments}

\subsection{Dataset and Evaluation Metric} 
Extensive experiments are conducted on five publicly available pedestrian attribute recognition datasets, including  \textbf{PETA}~\cite{deng2014peta}, \textbf{PA100K}~\cite{2017pa100k}, \textbf{RAPv1}~\cite{2016rapv1}, \textbf{RAPv2}~\cite{2019rapv2}, \textbf{WIDER}~\cite{li2016wider}. 
Additionally, we also compare existing methods on the facial attribute recognition (FAR) dataset \textbf{LFWA}~\cite{liu2015deep}.
In our experiments, five widely used metrics are adopted for the evaluation of PAR models, including \textbf{mA}, \textbf{Acc}, \textbf{Precision}, \textbf{Recall}, and \textbf{F1-score}. More details about the datasets and evaluation metrics can be found in our supplementary materials.

\begin{table*}[!htb]
\center
\caption{Comparison with state-of-the-art methods on PETA and PA100K datasets. The \textcolor{red}{first} and \textcolor{blue}{second} are shown in \textcolor{red}{red} and \textcolor{blue}{blue}, respectively. "-" means this indicator is not available. VTB* indicates that VTB uses CLIP's feature extractor.} \label{Comparisononpetadatasets} 
\begin{tabular}{l|c|ccccc|ccccc}
\hline \toprule [0.5 pt] 
\multicolumn{1}{c|}{\multirow{2}{*}{Methods}} & \multicolumn{1}{c|}{\multirow{2}{*}{Backbone}} & \multicolumn{5}{c|}{PETA} & \multicolumn{5}{c}{PA100K} \\ \cline{3-12} 
\multicolumn{1}{c|}{} &
  \multicolumn{1}{c|}{} &
  \multicolumn{1}{c}{mA} &
  \multicolumn{1}{c}{Acc} & 
  \multicolumn{1}{c}{Prec} &
  \multicolumn{1}{c}{Recall} &
  \multicolumn{1}{c|}{F1} &
  \multicolumn{1}{c}{mA} &
  \multicolumn{1}{c}{Acc} &
  \multicolumn{1}{c}{Prec} &
  \multicolumn{1}{c}{Recall} &
  \multicolumn{1}{c}{F1} \\ \hline
DeepMAR (ACPR 2015) \cite{deepmar} & CaffeNet & 82.89 & 75.07 & 83.68 & 83.14 & 83.41 & 72.70 & 70.39 & 82.24 & 80.42 & 81.32 \\
HPNet (ICCV 2017) \cite{2017pa100k} & Inception & 81.77 & 76.13 & 84.92 & 83.24 & 84.07 & 74.21 & 72.19 & 82.97 & 82.09 & 82.53 \\
JRL (ICCV 2017) \cite{wang2017JRL} & AlexNet & 82.13 & - & 82.55 & 82.12 & 82.02  & - & - & - & - & - \\
GRL (IJCAI 2018) \cite{zhao2018GRL} & Inception-V3 & 86.70 & - & 84.34 & 88.82 & 86.51  & - & - & - & - & - \\
MsVAA (ECCV 2018) \cite{2018msvaa} & ResNet101 & 84.59 & 78.56 & 86.79 & 86.12 & 86.46  & - & - & - & - & - \\
VAC (CVPR 2019) \cite{2019VAC} & ResNet50  & - & - & - & - & - & 79.16 & 79.44 & 88.97 & 86.26 & 87.59 \\
ALM (ICCV 2019) \cite{2019alm} & BN-Inception & 86.30 &  79.52 & 85.65 & 88.09 & 86.85 & 80.68 & 77.08 & 84.24 & 88.84 & 86.46 \\
JLAC (AAAI 2020) \cite{2020JLAC} & ResNet50 & 86.96 & 80.38 & 87.81 & 87.09 & 87.50 & 82.31 & 79.47 & 87.45 & 87.77 & 87.61 \\
SCRL (TCSVT 2020) \cite{wu2020person} & ResNet50 & 87.2 & -  & \textcolor{red}{\textbf{89.20}} & 87.5 & 88.3 & 80.6 & - & 88.7 & 84.9 & 82.1  \\
SSCsoft (ICCV 2021) \cite{2021ssc} & ResNet50 & 86.52 & 78.95 & 86.02 & 87.12 & 86.99 & 81.87 & 78.89 & 85.98 & 89.10 & 86.87 \\				
IAA-Caps (PR 2022) \cite{2022iaacaps} & OSNet & 85.27 & 78.04 & 86.08 & 85.80 & 85.64 & 81.94 & 80.31 & 88.36 & 88.01 & 87.80 \\
MCFL (NCA 2022) \cite{Chen2022MCFL} & ResNet-50 & 86.83 & 78.89 & 84.57 & 88.84 & 86.65 & 81.53 & 77.80 & 85.11 & 88.20 & 86.62 \\
DRFormer (NC 2022) \cite{2022drformer} & ViT-B/16 & \textcolor{red}{\textbf{89.96}} & \textcolor{blue}{\textbf{81.30}} & 85.68 & \textcolor{red}{\textbf{91.08}} & \textcolor{blue}{\textbf{88.30}} & 82.47 & 80.27 & 87.60 & 88.49 & 88.04 \\
VAC-Combine (IJCV 2022) \cite{guo2022visual} & ResNet50  & - & - & - & - & - & 82.19 & 80.66 & 88.72 & 88.10 & 88.41 \\
DAFL (AAAI 2022) \cite{jia2022learning} & ResNet50 & 87.07 & 78.88 & 85.78 & 87.03 & 86.40 & 83.54 & 80.13 & 87.01 & 89.19 & 88.09 \\
CGCN (TMM 2022) \cite{Fan2022CGCN} & ResNet & 87.08 & 79.30 & 83.97 & 89.38 & 86.59 & - & - & - & - & - \\ 
CAS-SAL-FR (IJCV 2022) \cite{yang2021cascaded} & ResNet50 & 86.40 & 79.93 & 87.03 & 87.33 & 87.18 & 82.86 & 79.64 & 86.81 & 87.79 & 85.18\\ 

VTB (TCSVT 2022) \cite{cheng2022VTB} & ViT-B/16 & 85.31 & 79.60 & 86.76 & 87.17 & 86.71 & 83.72 & 80.89 & 87.88 & 89.30 & 88.21 \\

\hline
VTB* (TCSVT 2022) \cite{cheng2022VTB} & ViT-L/14 & 86.34 & 79.59 & 86.66 & 87.82 & 86.97 & \textcolor{blue}{\textbf{85.30}} & \textcolor{blue}{\textbf{81.76}} & 87.87 & \textcolor{blue}{\textbf{90.67}} & \textcolor{blue}{\textbf{88.86}} \\	
PromptPAR (Ours) & ViT-L/14 & \textcolor{blue}{\textbf{88.76}} & \textcolor{red}{\textbf{82.84}} & \textcolor{blue}{\textbf{89.04}} & \textcolor{blue}{\textbf{89.74}} & \textcolor{red}{\textbf{89.18}} & \textcolor{red}{\textbf{87.47}} & \textcolor{red}{\textbf{83.78}} & \textcolor{blue}{\textbf{89.27}} & \textcolor{red}{\textbf{91.70}} & \textcolor{red}{\textbf{90.15}} \\
\hline \toprule [0.5 pt] 
\end{tabular}
\end{table*}

\subsection{Implementation Details} 

The input of visual encoder of CLIP model is $224 \times 224$, meanwhile, the pedestrian images are usually slender in shape. If we directly resize the given image into the square shape, it will significantly change the shape of the pedestrian and achieve poor recognition performance. Therefore, we process all the used PAR datasets by padding the pedestrian image using black pixels and changing the resolution to $224 \times 224$. Randomly cropping and horizontal flipping are utilized for data augmentation in the training phase. The ViT-L/14 version of CLIP~\cite{radford2021CLIP} is adopted as our visual encoder, the textual encoder contains 12 Transformer layers with 512 hidden nodes and 8 heads.  The number of prompt tokens for visual and textual encoders is 50 and 3, respectively. All these settings are shared for the experiments on RAPv1~\cite{2016rapv1}, RAPv2~\cite{2019rapv2}, PETA~\cite{deng2014peta}, and PA100K~\cite{2017pa100k} datasets.

We optimize our framework for 40 epochs using SGD optimizer~\cite{sutskever2013sgd} and set the initial learning rate of the classification head of MM-Former as 8e-3, and the initial learning rate of the prompt as 4e-3. Smaller learning rates (1e-3 and 2e-3) are set on the WIDER~\cite{li2016wider}, PETA-ZS~\cite{2021Rethinking}, and RAP-ZS~\cite{2021Rethinking} datasets for the classification head and prompt tokens respectively to achieve higher performances. We set the warm-up procedure for 5 epochs based on the cosine learning rate scheduler. The initial learning rate is decreased by a ratio of 0.01 during the warm-up, and the weight decay is 1e-4. The batch size is set as 16. The source code of this paper is released to help other researchers reproduce our experiments.

\begin{table*}
\center
\caption{Comparison with state-of-the-art methods on RAPv1 and RAPv2 datasets. The \textcolor{red}{first} and \textcolor{blue}{second} are shown in \textcolor{red}{red} and \textcolor{blue}{blue}, respectively. "-" means this indicator is not available.} \label{Comparisononrapdatasets} 
\begin{tabular}{l|c|ccccc|ccccc}
\hline \toprule [0.5 pt]
\multicolumn{1}{c|}{\multirow{2}{*}{Methods}} & \multicolumn{1}{c|}{\multirow{2}{*}{Backbone}} & \multicolumn{5}{c|}{RAPv1} & \multicolumn{5}{c}{RAPv2} \\ \cline{3-12} 
\multicolumn{1}{c|}{} &
  \multicolumn{1}{c|}{} &
  \multicolumn{1}{c}{mA} &
  \multicolumn{1}{c}{Acc} &
  \multicolumn{1}{c}{Prec} &
  \multicolumn{1}{c}{Recall} &
  \multicolumn{1}{c|}{F1} &
  \multicolumn{1}{c}{mA} &
  \multicolumn{1}{c}{Acc} &
  \multicolumn{1}{c}{Prec} &
  \multicolumn{1}{c}{Recall} &
  \multicolumn{1}{c}{F1} \\ \hline
    DeepMAR (ACPR 2015) \cite{deepmar} & CaffeNet & 73.79 & 62.02 & 74.92 & 76.21 & 75.56 & - & - & - & - & - \\
    HPNet (ICCV 2017) \cite{2017pa100k} & Inception & 76.12 & 65.39 & 77.33 & 78.79 & 78.05 & - & - & - & - & - \\
    JRL (ICCV 2017) \cite{wang2017JRL} & AlexNet & 74.74 & - & 75.08 & 74.96 & 74.62 & - & - & - & - & - \\
    GRL (IJCAI 2018) \cite{zhao2018GRL} & Inception-V3 & 81.20 & - & 77.70 & 80.90 & 79.29 & - & - & - & - & - \\
    MsVAA (ECCV 2018) \cite{2018msvaa} & ResNet101 & - & - & - & - & - & 78.34 & 65.57 & \textcolor{blue}{\textbf{77.37}} & 79.17 & 78.26  \\
    VAC (CVPR 2019) \cite{2019VAC} & ResNet50 & - & - & - & - & - & 79.23 & 64.51 & 75.77 & 79.43 & 77.10 \\
    ALM (ICCV 2019) \cite{2019alm} & BN-Inception & 81.87 & 68.17 & 74.71 & 86.48 & 80.16 & 79.79 & 64.79 & 73.93 & 82.03 & 77.77 \\
    JLAC (AAAI 2020) \cite{2020JLAC} & ResNet50 & 83.69 & 69.15 & 79.31 & 82.40 & 80.82 & 79.23 & 64.42 & 75.69 & 79.18 & 77.40 \\
    SSCsoft (ICCV 2021) \cite{2021ssc} & ResNet50 & 82.77 & 68.37 & 75.05 & 87.49 & 80.43 
    & - & - & - & - & - \\
    IAA-Caps (PR 2022) \cite{2022iaacaps} & OSNet & 81.72 & 68.47 & 79.56 & 82.06 & 80.37 & - & - & - & - & - \\
    MCFL (NCA 2022) \cite{Chen2022MCFL} & ResNet50 & 84.04 & 67.28 & 73.44 & \textcolor{blue}{\textbf{87.75}} & 79.96 & - & - & - & - & - \\
    DRFormer (NC 2022) \cite{2022drformer} & ViT-B/16 & 81.81 & \textcolor{blue}{\textbf{70.60}} & 80.12 & 82.77 & 81.42 & - & - & - & - & - \\
    VAC-Combine (IJCV 2022) \cite{guo2022visual} & ResNet50 & 81.30 & 70.12 & \textcolor{blue}{\textbf{81.56}} & 81.51 & \textcolor{blue}{\textbf{81.54}} 
    & - & - & - & - & - \\
    DAFL (AAAI 2022) \cite{jia2022learning} & ResNet50 & 83.72 & 68.18 & 77.41 & 83.39 & 80.29 & 81.04 & 66.70 & 76.39 & 82.07 & 79.13 \\
    CGCN (TMM 2022) \cite{Fan2022CGCN} & ResNet50 & \textcolor{blue}{\textbf{84.70}} & 54.40 & 60.03 & 83.68 & 70.49 & - & - & - & - & -  \\ 
    CAS-SAL-FR (IJCV 2022) \cite{yang2021cascaded} & ResNet50 & 84.18 & 68.59 & 77.56 & 83.81 & 80.56 & - & - & - & - & - \\ 
    
    VTB (TCSVT 2022) \cite{cheng2022VTB} & ViT-B/16 & 82.67 & 69.44 & 78.28 & 84.39 & 80.84 & 81.34 
    & 67.48 & 76.41 & 83.32 & 79.35 \\
    PARformer (TCSVT 2023) \cite{fan2023parformer} & Swin-L & 84.13 & 69.94 & 79.63 & \textcolor{red}{\textbf{88.19}} & 81.35 & - 
    & - & -& -  & - \\
    \hline
    VTB* (TCSVT 2022) \cite{cheng2022VTB} & ViT-L/14 & 83.69 & 69.78 & 78.09 & 85.21 & 81.10 & \textcolor{blue}{\textbf{81.36}} & \textcolor{blue}{\textbf{67.58}} & 76.19 & \textcolor{blue}{\textbf{84.00}} & \textcolor{blue}{\textbf{79.52}} \\	
    PromptPAR (Ours)  & ViT-L/14 & \textcolor{red}{\textbf{85.45}} & \textcolor{red}{\textbf{71.61}} & 79.64 & 86.05 & \textcolor{red}{\textbf{82.38}} 
    & \textcolor{red}{\textbf{83.14}} & \textcolor{red}{\textbf{69.62}} & \textcolor{red}{\textbf{77.42}} & \textcolor{red}{\textbf{85.73}} & \textcolor{red}{\textbf{81.00}} \\
    \hline \toprule [0.5 pt] 
\end{tabular}
\end{table*}

\subsection{Compare with Other SOTA Algorithms}

\begin{figure}  
\centering
\small
\includegraphics[width=0.45\textwidth]{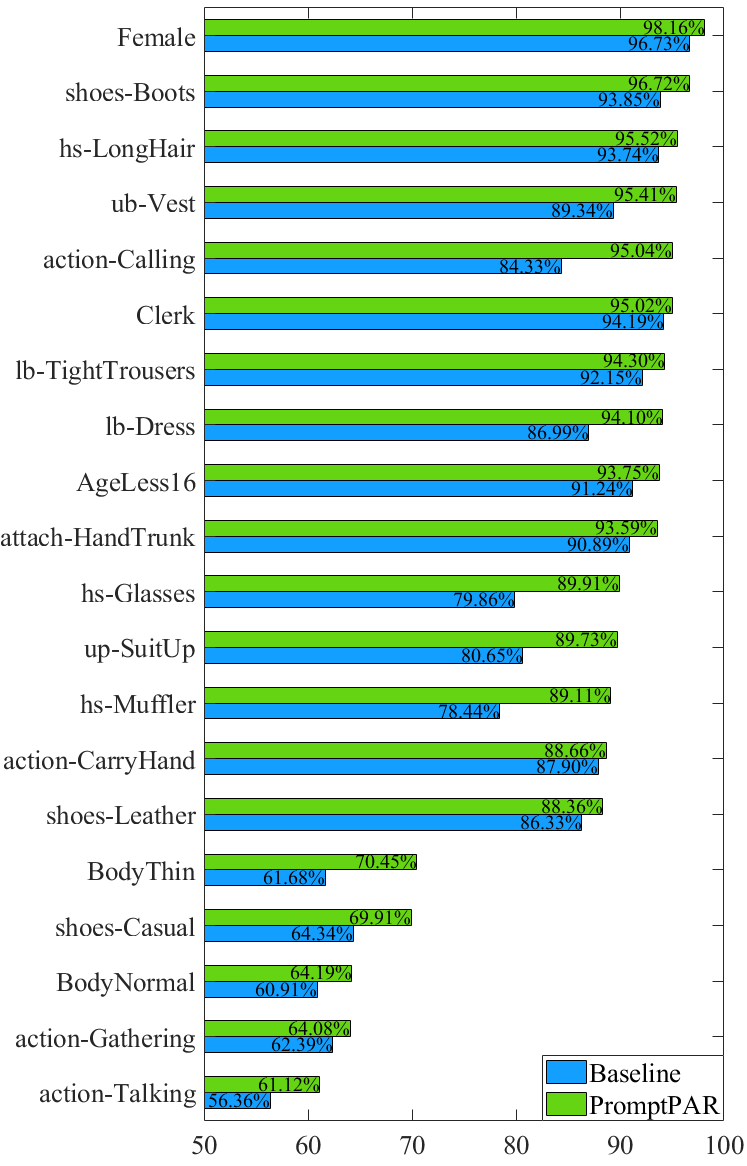}
\caption{Average precision of 20 pedestrian attributes on the RAP-V1 dataset.} 
\label{meanAcc20attributes}
\end{figure}

\textbf{Results on PA100K~\cite{2017pa100k} Dataset.~} 
As shown in Table~\ref{Comparisononpetadatasets}, our baseline method VTB (ViT-B/16 is adopted as the backbone) achieves 83.72, 80.89, 87.88, 89.30, 88.21 on the mA, Accuracy, Precision, Recall, and F1-score, which is already better than other state-of-the-art algorithms, such as DAFL~\cite{jia2022learning} (AAAI 2022), DRFormer~\cite{2022drformer} (NC 2022), SSCsoft~\cite{2021ssc} (ICCV 2021). In contrast, our proposed PromptPAR achieves 87.47, 83.78, 89.27, 91.70, 90.15 on the five metrics respectively, which improves the baseline by a large margin, i.e., +3.75, +2.89, +1.39, +2.40, +1.94. 
We also integrate the ViT-L/14 based CLIP visual encoder into the VTB algorithm and it achieves better results than the ViT-B/16 based version (i.e., 86.34, 79.59, 86.66, 87.82, 86.97 on these metrics), however still inferior to ours. 
Our method also beats all the compared methods on the most of evaluation metrics, even the strong PAR algorithm VAC-Combine~\cite{guo2022visual} (IJCV 2022).

\textbf{Results on PETA~\cite{deng2014peta} Dataset.~} 
As shown in Table~\ref{Comparisononpetadatasets}, we train the VTB~\cite{cheng2022VTB} on the PETA dataset using their default settings and compare them with ours. Specifically, the baseline VTB (ViT-B/16 version) achieves 85.31, 79.60, 86.76, 87.17, 86.71 on the mA, Accuracy, Precision, Recall, and F1-score, which are comparable with many existing works. In contrast, our proposed PromptPAR model obtains 88.76, 82.84, 89.04, 89.74, 89.18 on these metrics and improves the baseline significantly. We also beat many SOTA pedestrian attribute recognition models, such as CAS-SAL-FR (IJCV 2022)~\cite{yang2021cascaded}, IAA-Caps (PR 2022) \cite{2022iaacaps}, SSCsoft (ICCV 2021)~\cite{2021ssc}, etc. These experiments fully validated the effectiveness and advantages of our proposed modules for pedestrian attribute recognition. Note that, our model also achieves comparable performance compared with DRFormer (NC 2022)~\cite{2022drformer}, which also adopts ViT-B/16 as the backbone. However, their parameters of Transformer encoder are initialized using the weights of ViT-B/16 pre-trained on ImageNet dataset~\cite{deng2009imagenet}, meanwhile, the CLIP model used in our framework is pre-trained on image-language pairs which are easily collected and annotation free. In addition, our model achieves better results than DRFormer~\cite{2022drformer} on other PAR datasets which will be introduced next.

\textbf{Results on RAPv1~\cite{2016rapv1} \& RAPv2~\cite{2019rapv2} Dataset.~} 
As shown in Table~\ref{Comparisononrapdatasets}, our baseline method VTB (ViT-B/16 version) achieves 82.67, 69.44, 78.28, 84.39, 80.84 on the mA, Accuracy, Precision, Recall, and F1-score, which are already better than most state-of-the-art algorithms, such as VAC-Combine~\cite{guo2022visual} (IJCV2022), DRFormer~\cite{2022drformer} (NC 2022), IAA-Caps~\cite{2022iaacaps} (PR 2022). Compared to VTB, our PromptPAR achieves new SOTA results on three main metrics, i.e., 85.96, 71.95, 81.98 on the mA, Accuracy, and F1-score.

In addition to the overall results on the RAP-V1 dataset, we also report the recognition performance of some specific attributes, as shown in Fig.~\ref{meanAcc20attributes}. The blue and green rectangles are the results of the baseline approach and our proposed PromptPAR. It is easy to find that our method achieves better results on most of the compared attributes, for example, we achieve 89.91\% on \emph{hs-Glasses}, and the baseline attains 79.86\%. Our results are also better than the baseline on the action attributes, such as \emph{action-Calling, action-CarryingHand}, and \emph{action-Gathering}. These experimental results fully demonstrate the effectiveness of our proposed modules for the PAR task. Similar conclusions can be drawn from the results on the RAPv1~\cite{2016rapv1} dataset.

\begin{table}
\center
\small  
\caption{Comparison with state-of-the-art methods on WIDER datasets. } \label{Comparisononwiderdatasets} 
\begin{tabular}{l|c|c}
\hline \toprule [0.5 pt]
\multicolumn{1}{c|}{\multirow{1}{*}{Methods}} & \multicolumn{1}{c|}{\multirow{1}{*}{Backbone}} & \multicolumn{1}{c}{mA} \\
  \hline
    R*CNN (ICCV 2015)\cite{2015rcnn} & VGG16 & 80.5\\
    DHC (ECCV 2016)\cite{li2016wider} & VGG16 & 81.3\\
    SRN (CVPR 2017)\cite{2017srn} & ResNet101 & 86.2\\
    DIAA (ECCV 2018)\cite{2018diaa} & ResNet101 & 86.4\\
    Da-HAR (AAAI 2020)\cite{2019dahar} & ResNet101 & 87.3\\
    VAC-Combine (IJCV 2022) \cite{guo2022visual} & ResNet50 & 88.4\\
    
    VTB (TCSVT 2022) \cite{cheng2022VTB} & ViT-B/16 & 88.2\\
    \hline
    VTB* (TCSVT 2022) \cite{cheng2022VTB} & ViT-L/14 & \textcolor{blue}{\textbf{91.2}}\\
    PromptPAR (Ours)  & ViT-L/14 &  \textcolor{red}{\textbf{92.0}}\\  
    \hline \toprule [0.5 pt] 

\end{tabular} 
\end{table}

\begin{table}
\center
\small  
\caption{Comparison with state-of-the-art methods on LFWA~\cite{liu2015deep} datasets.(measured by mA (\%)} \label{ComparisononLFWAdatasets} 
\begin{tabular}{l|c}
\hline \toprule [0.5 pt]
\multicolumn{1}{c|}{Methods}  & \multicolumn{1}{c}{LFWA} \\ \hline
SSP+SSG~\cite{kalayeh2017improving} & \textcolor{blue}{\textbf{87.13}} \\
MCNN-AUX~\cite{hand2017mcnn} & 86.31 \\
He et al.~\cite{he2018harnessing} & 85.28 \\
AFFAIR~\cite{li2018landmark} & 86.13 \\
GNAS~\cite{huang2018gnas} & 86.37 \\
DMM-CNN~\cite{mao2020dmm} & 86.56\\
SSPL~\cite{shu2021sspl} & 86.53  \\ 
ICKD~\cite{chen2023ickd} & 86.91 \\ \hline
VTB(Baseline)~\cite{cheng2022VTB}  & 85.35 \\ 
PromptPAR (Ours)  & \textcolor{red}{\textbf{87.16}}  \\  
    \hline \toprule [0.5 pt] 

\end{tabular} 
\end{table}

\textbf{Results on WIDER~\cite{li2016wider} Dataset.~}
As shown in Table~\ref{Comparisononwiderdatasets}, we refine our baseline method VTB on the WIDER dataset and achieve 88.2 on the mA, which already surpasses most previous algorithms, such as DIAA~\cite{2018diaa} (ECCV 2018), Da-HAR~\cite{2018diaa} (AAAI 2020). Our method has improved considerably compared to it, reaching 92.0, an improvement of +3.8. Our method also beats all the compared methods on the most of evaluation metrics, even the strong PAR algorithm VAC-Combine~\cite{guo2022visual} (IJCV 2022).

\textbf{Results on FAR Dataset.~} 
 The LFWA~\cite{liu2015deep} FAR dataset, widely used in facial attribute recognition, consists of 6,263 face images in the training set and 6,880 in the test set, with 40 attributes annotated for each face image. Our method performs better than existing facial attribute recognition methods, with a 1.81\% improvement in mA over our baseline method and achieving approximately 0.25\% improvement in mA compared to ICKD (TCSVT2023)~\cite{chen2023ickd}.

All in all, we can find that our proposed PromptPAR achieves better results than most of the compared PAR models, or even set new SOTA performance on the RAP-V1, RAP-V2, PA100K, and WIDER datasets. The strong performance on multiple datasets fully validated the effectiveness of our proposed multi-modal pre-trained models and prompt tuning for the pedestrian attribute recognition task. 

\begin{table}
\center
\small  
\caption{Comparison different CLIP model.} 
\label{Backbone} 
\resizebox{\columnwidth}{!}{ 
\begin{tabular}{c|c|c|c|cc}
\hline \toprule [0.5 pt]
\multicolumn{1}{c|}{\multirow{2}{*}{Method}}  &\multicolumn{1}{c|}{\multirow{2}{*}{Backbone}} &\multicolumn{1}{c|}{\multirow{2}{*}{Source}} &\multicolumn{1}{c|}{\multirow{2}{*}{Params(M)}} &  \multicolumn{2}{c}{RAPv1}  \\ \cline{5-6} 
\multicolumn{1}{c|}{} & 
\multicolumn{1}{c|}{} &
\multicolumn{1}{c|}{} &
\multicolumn{1}{c|}{} &
\multicolumn{1}{c}{mA} &
\multicolumn{1}{c}{F1} \\ 
\hline
 VTB \cite{cheng2022VTB} & ViT-B/16 &ImageNet & 87.15 & 82.67 & 80.84  \\
 DFDT \cite{ZHENG2023105708} & Swin-B  &ImageNet & 87.59 &  82.34 & 82.15
 \\ \hline
 VTB \cite{cheng2022VTB} & ViT-B/16 &CLIP     & 157.54 & 83.26 & 81.05 \\
 Ours & ViT-B/16 &CLIP    & \textbf{2.30} & 83.74 & 81.07 \\
 \hline
 VTB \cite{cheng2022VTB} & ViT-L/14 &CLIP    & 435.93 & 85.11 & 82.02 \\
 Ours & ViT-L/14 &CLIP    & \textbf{3.27} & 85.45 & 82.38  \\
\hline \toprule [0.5 pt] 
\end{tabular} } 
\end{table}

\begin{table}
\center
\small  
\caption{Comparison of different MM-Former layers. } \label{MMformer} 
\begin{tabular}{l|cc|cc}
\hline \toprule [0.5 pt]
\multicolumn{1}{c|}{\multirow{2}{*}{MM-Former}} &  \multicolumn{2}{c|}{RAPv1} &  \multicolumn{2}{c}{PETA} \\ \cline{2-5} 
\multicolumn{1}{c|}{} &
\multicolumn{1}{c}{mA} &
\multicolumn{1}{c|}{F1} &
\multicolumn{1}{c}{mA} &
\multicolumn{1}{c}{F1} \\ \hline
MLP & 85.54 & 82.07 & 87.85 & 88.05 \\
MM-Former(1-layer) & 85.45 & 82.38 & 88.76 & 89.18 \\
MM-Former(2-layers) & 85.55 & 82.43 & 88.92 & 89.02 \\
MM-Former(4-layers) & 85.12 & 82.40 & 88.71 & 88.98 \\
\hline \toprule [0.5 pt] 
\end{tabular} 
\end{table}

\begin{table*}[!htp]
\center
\small      
\caption{Component Analysis on the RAP-v1 Dataset. mA, Acc, and F1 results are reported. The Expand Sen. denotes expanding the attributes to sentences as text prompt.} 
\label{CAResults} 
\resizebox{0.9\textwidth}{13mm}{
\begin{tabular}{c|c|c|c|c|c|c|c|ccc|ccc} 		
\hline \toprule [0.5 pt] 
\multicolumn{1}{c|}{\multirow{2}{*}{No.}} & \multicolumn{1}{c|}{\multirow{2}{*}{ViT}} & \multicolumn{1}{c|}{\multirow{2}{*}{PTM}} & \multicolumn{1}{c|}{\multirow{2}{*}{FTune}} & \multicolumn{1}{c|}{\multirow{2}{*}{PTune(V)}} & \multicolumn{1}{c|}{\multirow{2}{*}{PTune(T)}} & \multicolumn{1}{c|}{\multirow{2}{*}{RegionPTune}}  & \multicolumn{1}{c|}{\multirow{2}{*}{Expand Sen.}}  & \multicolumn{3}{c|}{RAPv1} & \multicolumn{3}{c}{PETA} \\ \cline{9-14} 
  \multicolumn{1}{c|}{} &
  \multicolumn{1}{c|}{} &
  \multicolumn{1}{c|}{} &
    \multicolumn{1}{c|}{} &
  \multicolumn{1}{c|}{} &
  \multicolumn{1}{c|}{} &
  \multicolumn{1}{c|}{} &
  \multicolumn{1}{c|}{} &
  \multicolumn{1}{c}{mA} &
  \multicolumn{1}{c}{Acc} &
  \multicolumn{1}{c|}{F1}  & 
  \multicolumn{1}{c}{mA} &
  \multicolumn{1}{c}{Acc} &
  \multicolumn{1}{c}{F1} 
\\ 
\hline 
1  &\cmark &       &\cmark &\cmark &       &       &       & 82.79 & 68.95 & 80.43 & 85.31 & 79.60 & 86.71 \\ \hline
2  &       &\cmark &       &\cmark &       &       &       & 85.22 & 71.12 & 82.12 & 88.54 & 82.25 & 88.77 \\ \hline
3  &       &\cmark &       &\cmark &\cmark &       &       & 85.37 & 71.27 & 82.20 & 88.54 & 82.46 & 88.94 \\ \hline
4  &       &\cmark &       &\cmark &\cmark &\cmark &       & 85.40 & 71.46 & 82.33 & 88.63 & 82.47 & 88.96 \\ \hline
5  &       &\cmark &       &\cmark &\cmark &\cmark &\cmark & 85.45 & 71.61 & 82.38 & 88.76 & 82.82 & 89.18 \\ 
\hline \toprule [0.5 pt]
\end{tabular}}
\end{table*} 

\begin{table}
\center
\small  
\caption{Comparison of different prompt initialization methods. } \label{initialization} 
\begin{tabular}{l|cc}
\hline \toprule [0.5 pt]
\multicolumn{1}{c|}{\multirow{2}{*}{Initialization}} &  \multicolumn{2}{c}{RAPv1}  \\ \cline{2-3} 
\multicolumn{1}{c|}{} &
\multicolumn{1}{c}{mA} &
\multicolumn{1}{c}{F1} \\ \hline
Random-init & 85.45 & \textbf{82.38} \\
Zero-Init & \textbf{85.81} & 82.23 \\
\hline \toprule [0.5 pt] 
\end{tabular} 
\end{table}

\subsection{Ablation Study} 
In this subsection, we conduct extensive experiments to validate the effectiveness of the key modules in our proposed PromptPAR framework. Specifically, the backbone network, fuse module, and optimization methods are all discussed, as shown in {Table~\ref{Backbone},~\ref{MMformer} and ~\ref{CAResults}. More details will be introduced in the following paragraphs respectively.

\textbf{Effects of Pre-trained Multi-modal Models.~} In this work, we employed CLIP as the Multi-modal Pre-trained Big Model (MM-PTM) for extracting visual and text features. In contrast to independently trained visual or text models, such as ViT~\cite{dosovitskiy2020VIT} and BERT~\cite{kenton2019bert}, as used in our baseline approach VTB~\cite{cheng2022VTB}, the MM-PTM provides us with more robust feature representations. To illustrate, as depicted in Table~\ref{Backbone}, VTB~\cite{cheng2022VTB}, utilizing the ViT-B/16~\cite{dosovitskiy2020VIT} and BERT~\cite{kenton2019bert} models, achieves 82.67 and 80.84 in mean accuracy (mA) and F1 score, respectively. When we replaced these two backbones with CLIP ViT-B/16 and ViT-L/14, the overall performance improved to 83.26, 81.05, and 85.11, 82.02, respectively. This experiment underscores the effectiveness of MM-PTM in the context of PAR tasks.
It is also worth noting that the ViT~\cite{dosovitskiy2020VIT} model used in existing PAR works~\cite{cheng2022VTB, 2022drformer} is initialized with the weights pre-trained on the ImageNet dataset. Although good performance can be obtained, however, their results are actually built on expensive annotation. In contrast, the used CLIP model in our framework is pre-trained on vision-language pairs in an unsupervised way. Thus, it will be an interesting and economical way to learn the PAR model based on pre-trained big models. 

\textbf{Finetuning vs Prompt Tuning.~}In this work, we are inspired by the VPT~\cite{2022vpt} which optimizes only very few parameters of our model. As shown in Table~\ref{Backbone}, the fine-tuning based model achieves 85.11, 82.02 on mA and F1 score. When replacing the fine-tuning with our proposed vision-language prompt tuning, the overall performance can be improved to 85.45 and 82.38. Due to the challenges associated with Full Fine-tuning of CLIP, ensuring optimal performance is quite challenging. As a result, the prompt tuning approach we introduce here offers slightly superior performance compared to Full Fine-tuning. Notably, when utilizing the ViT-B/16 and ViT-L/14 models, prompt tuning only requires approximately 1.46\% and 0.75\% of the full fine-tuned parameters. This observation underscores the efficacy of prompt tuning.

\textbf{Effects of Multi-modal Transformers.~} As shown in Table~\ref{MMformer}, we conducted a comparison between MLPs and Multi-modal Transformers with varying numbers of layers to assess the efficacy of Multi-modal Transformers. When we replaced a single layer MM-Former with a fully connected layer, the performance changes on RAPv1 and PETA datasets were +0.09 and -0.31, as well as -0.91 and -1.13, respectively. These results strongly demonstrate the effectiveness of Multi-modal Transformers. It is evident that merely augmenting the number of layers in the MM-Former does not yield performance improvements. In fact, the increase in the number of layers results in a higher computational load. Consequently, in this paper, we employ only a single layer for our MM-Former.

\textbf{Effects of Visual and Text Prompt.~} As validated in the previous subsection, prompt learning plays an important role in our PromptPAR framework. In this part, we conduct a more fine-grained experiment to check the influence of visual/text prompt separately. As shown in Table~\ref{CAResults}, the visual prompt-based model, i.e., Algorithm 2, achieves $85.22, 71.12, 82.12$ on mA, Acc, and F1. We can also find that the overall performance can be improved by introducing the text prompt from 85.22, 71.12, 82.12 to 85.37, 71.27, 82.20. We can also find that our newly proposed region-aware prompt tuning strategy resulted in an improvement of +0.08 in mean accuracy (ma), +0.34 in accuracy (acc), and +0.18 in F1 score, as compared to the previous global prompt. 
These results demonstrate the benefits of visual/text prompt tuning for attribute recognition.

\textbf{Effects of Different Prompt Initialization Methods.~} Inspired by LLaMA-Adapter~\cite{zhang2024llamaadapter}, we analyzed the initialization method for Region-aware Prompts, as shown in Table~\ref{initialization}. We employed two distinct initialization methods: zero initialization and random initialization for Region-aware Prompts, and we can find that the zero-initialized Region-aware Prompts exhibit a 0.36\% improvement in mA metrics and a 0.15\% reduction in F1 metrics, compared to the random initialization. The results demonstrate that the different initialization methods did not significantly affect the final outcomes.

\textbf{Effects of Using Expanded Sentences As Text Prompts.~} We are inspired by the prompt engineering of CLIP~\cite{radford2021CLIP} to expand attributes into sentences, as shown in Table~\ref{CAResults}, we conduct an ablation study using expanded sentences as text prompts. On the RAPv1 dataset, attribute expansion improves by 0.03, 0.15, and 0.05 in mA, Acc, and F1, respectively. The same enhancement is also found on the PETA dataset. The results demonstrate the validity of using expanded attributes.

\begin{table*}[!htb]
\center
\small  
\caption{Comparison with state-of-the-art methods on PETA-ZS and RAP-ZS datasets.} \label{PARZSresults} 
\begin{tabular}{l|c|ccccc|ccccc}
\hline \toprule [0.5 pt]
\multicolumn{1}{c|}{\multirow{2}{*}{Methods}} & \multicolumn{1}{c|}{\multirow{2}{*}{Backbone}} & \multicolumn{5}{c|}{PETA-ZS} & \multicolumn{5}{c}{RAP-ZS} \\ \cline{3-12} 
\multicolumn{1}{c|}{} &
  \multicolumn{1}{c|}{} &
  \multicolumn{1}{c}{mA} &
  \multicolumn{1}{c}{Acc} &
  \multicolumn{1}{c}{Prec} &
  \multicolumn{1}{c}{Recall} &
  \multicolumn{1}{c|}{F1} &
\multicolumn{1}{c}{mA} &
  \multicolumn{1}{c}{Acc} &
  \multicolumn{1}{c}{Prec} &
  \multicolumn{1}{c}{Recall} &
  \multicolumn{1}{c}{F1} 
\\ 
\hline
MsVAA (ECCV 2018) \cite{2018msvaa} & ResNet101 & 71.53 & 58.67 & 74.65 & 69.42 & 71.94 & 72.04 & 62.13 & 75.67 & 75.81 & 75.74\\
VAC (CVPR 2019) \cite{2019VAC} & ResNet50 & 71.91 & 57.72 & 72.05 & 70.64 & 70.90 & 73.70 & 63.25 & 76.23 & 76.97 & 76.12 \\
ALM (ICCV 2019) \cite{2019alm} & BN-Inception & 73.01 & 57.78 & 69.50 & 73.69 & 71.53  & 74.28 & 63.22 & 72.96 & 80.73 & 76.65 \\
JLAC (AAAI 2020) \cite{2020JLAC} & ResNet50 & 73.60 & 58.66 & 71.70 & 72.41 & 72.05 & 76.38 & 62.58 & 73.14 & 79.20 & 76.05 \\
Jia et al. (Arxiv 2021) \cite{2021Rethinking} & ResNet50 & 71.62 & 58.19 & 73.09 & 70.33 & 71.68 & 72.32 & 63.61 & \textcolor{blue}{\textbf{76.88}} & 76.62 & 76.75 \\
MCFL (NCA 2022) \cite{Chen2022MCFL} & ResNet50 & 72.91 & 57.04 & 68.47 & 74.35 & 71.29 & 74.37 & 63.37 & 71.21 & 83.86 & 77.02 \\

VTB (TCSVT 2022) \cite{cheng2022VTB} & ViT-B/16    & 75.13 & 60.50 & 73.29 & 74.40 & 73.38 & 75.76 & 64.73 & 74.93 & 80.85 & 77.35 \\
\hline
VTB* (TCSVT 2022) \cite{cheng2022VTB} & ViT-L/14    & \textcolor{blue}{\textbf{77.18}} & \textcolor{blue}{\textbf{63.12}} & \textcolor{blue}{\textbf{74.77}} & \textcolor{blue}{\textbf{77.24}} & \textcolor{blue}{\textbf{75.50}} & \textcolor{blue}{\textbf{79.17}} & \textcolor{blue}{\textbf{68.34}} &76.81 & \textcolor{blue}{\textbf{84.51}} & \textcolor{blue}{\textbf{80.07}} \\
PromptPAR (Ours)  & ViT-L/14                       & \textcolor{red}{\textbf{80.08}} & \textcolor{red}{\textbf{66.02}} & \textcolor{red}{\textbf{76.53}} & \textcolor{red}{\textbf{80.49}} & \textcolor{red}{\textbf{77.77}}         & \textcolor{red}{\textbf{80.43}} & \textcolor{red}{\textbf{70.39}} & \textcolor{red}{\textbf{78.48}} & \textcolor{red}{\textbf{85.57}} & \textcolor{red}{\textbf{81.52}}   \\
\hline \toprule [0.5 pt] 
\end{tabular} 
\end{table*}

\subsection{Analysis on Zero-Shot PAR}  
Extensive experiments conducted on the standard set of pedestrian attribute recognition fully validated the effectiveness of our PromptPAR. In this part, we also analyze our model on the zero-shot PAR which split the training and testing subset without overlapped person identity. Jia et al.~\cite{2021Rethinking} first propose such a setting to better fit the practical application scenarios. Following Jia et al.~\cite{2021Rethinking}, two datasets PETA-ZS~\cite{2021Rethinking} and RAP-ZS~\cite{2021Rethinking} are adopted and the corresponding results are reported in Table~\ref{PARZSresults}. We can find that our proposed PromptPAR achieves significant improvements on both datasets over the baseline approach and also other state-of-the-art PAR algorithms. To be specific, our baseline obtains 75.13, 60.50, and 73.38 on the mA, Accuracy, and F1 score on the PETA dataset, while we achieve 80.08, 66.02, and 77.77 on these metrics. The improvements are up to +4.95, +5.52, and +4.39, respectively. Similar results can be found based on the RAP-ZS dataset. All in all, these experimental results fully demonstrate the generalization and strong performance of our proposed modules for the PAR task.

\subsection{Parameter Analysis} 

In this part, we analyze the influence of different lengths and frequency/depth of visual prompts used in our model.

\textbf{Analysis on the Length of Prompts.~} 
As shown in Fig.~\ref{ParameterAnalysis} (a), we set the prompt length as 5, 10, 25, 50, 100, and 200, and test our model on the RAPv1 dataset. It is easy to find that the recognition results are relatively stable, more in detail, the accuracy, F1-score, and mA range from 71\%-72\%, 85\%-86\%, and 82-83\% on this dataset. This experiment demonstrates that our model is robust to the length of prompt vectors.

\begin{figure}
\centering
\small
\includegraphics[width=0.5\textwidth]{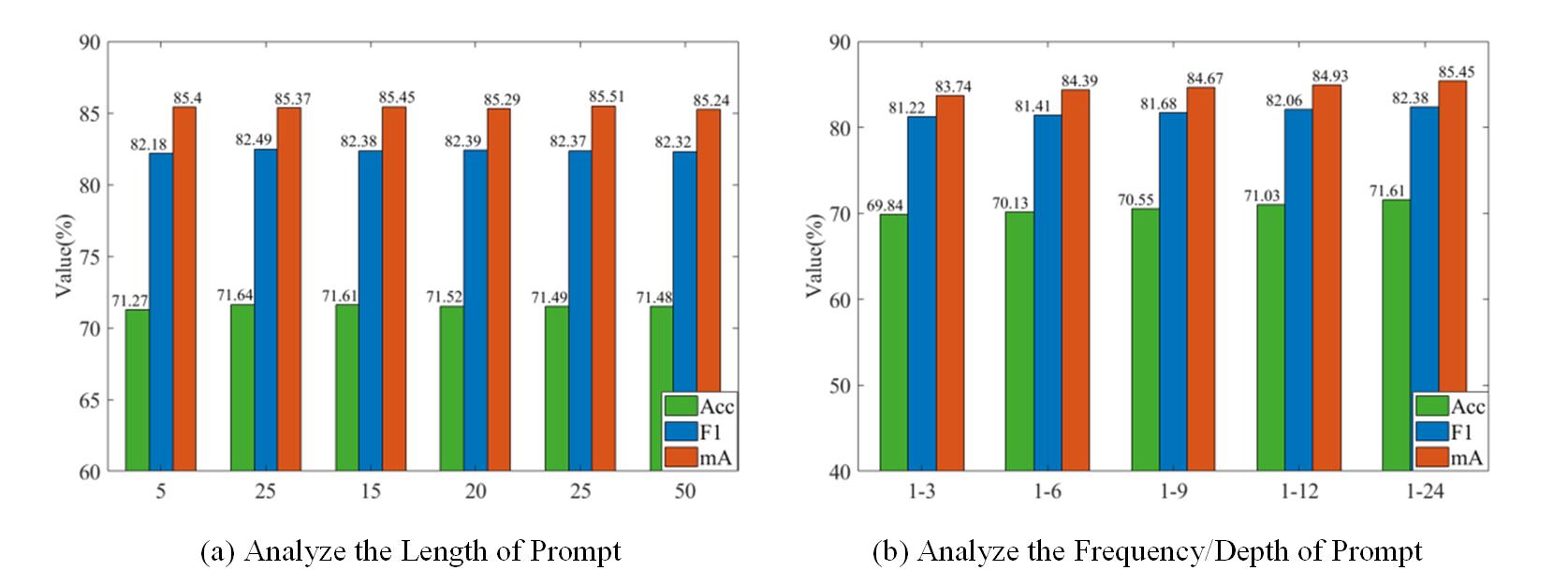}
\caption{Experimental results of different lengths and frequencies of prompt on RAP-v1 dataset.}
\label{ParameterAnalysis}
\end{figure}

\textbf{Analysis on the Frequency/Depth of Prompts.~}  
The prompt vectors can be injected into multiple layers of CLIP models as part of the information for message passing. As shown in Fig.~\ref{ParameterAnalysis} (b), we test our model with different frequencies or depths for PAR, including 3, 6, 9, 12, and 24 layers. The corresponding results are 69.84, 70.13, 70.55, 71.03, 71.61 on Accuracy, 81.22, 81.41, 81.68, 82.06, 82.38 on F1-score, and 83.74, 84.39, 84.67, 84.93, 85.45 on mA metric. We can find that better results can be achieved when the depth is set as 24.

\begin{figure}[!htp]
\centering
\small
\includegraphics[width=0.45\textwidth]{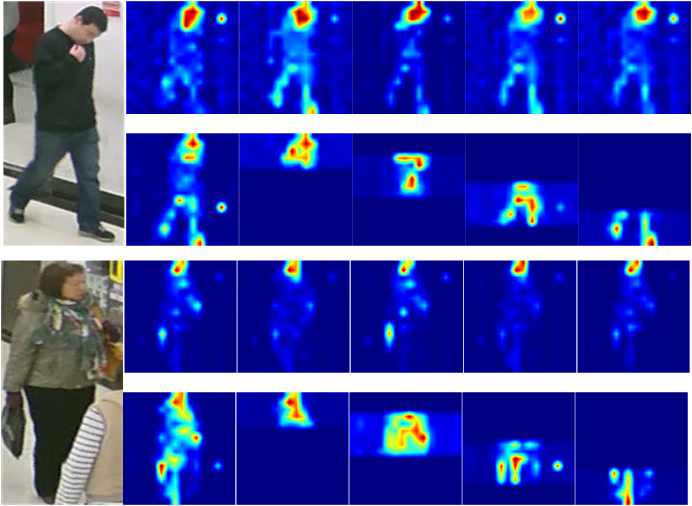}
\caption{Visualization of the response maps between prompts and pedestrian images. The \textcolor{red}{red} and \textcolor{blue}{blue} colors denote the high and low response. The top represents the response graph from the previous global prompt method, and the bottom represents the response graph from our RegionPTune}    
\label{respond} 
\end{figure} 

\begin{figure}[!htp]
\centering
\small
\includegraphics[width=0.5\textwidth]{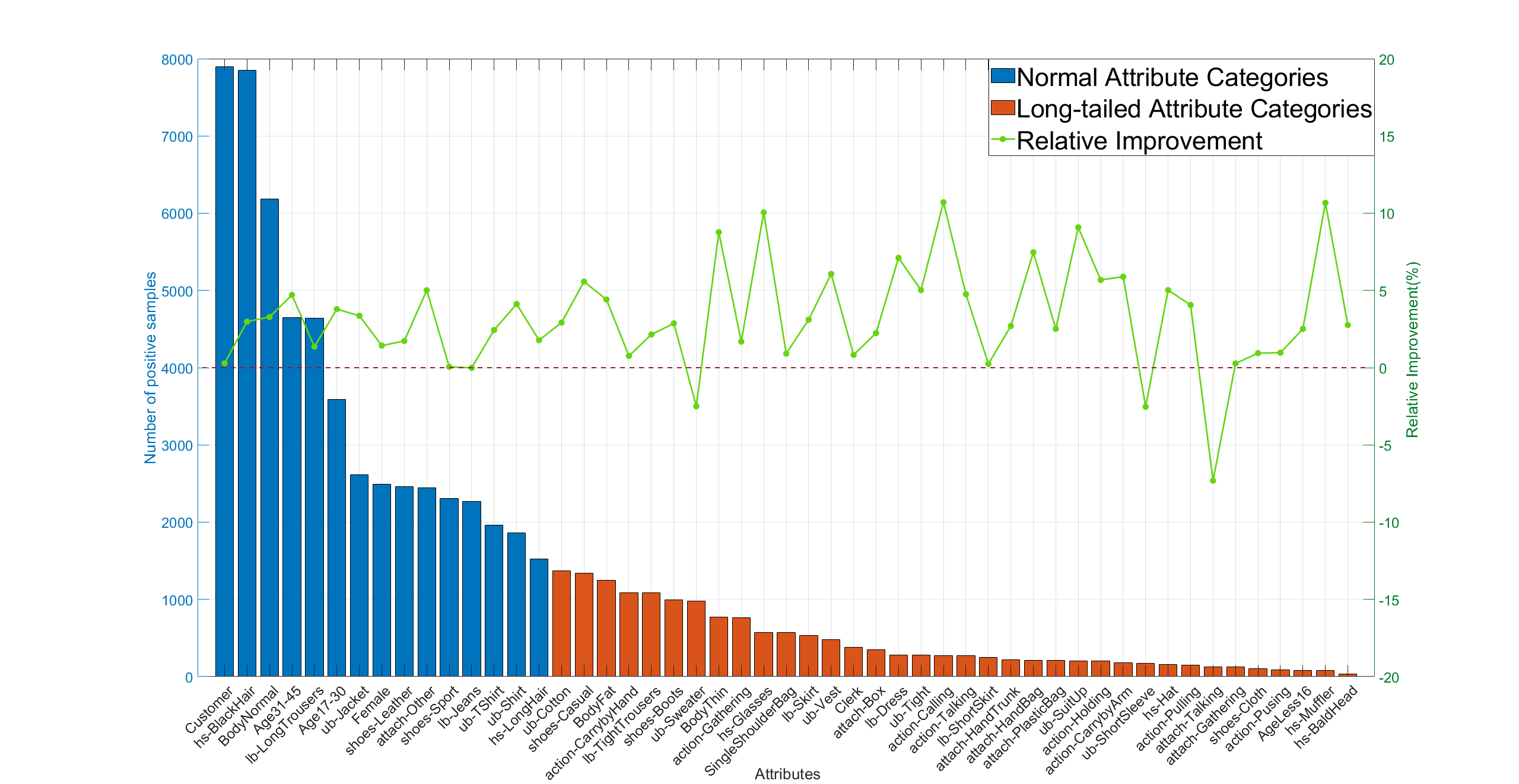}
\caption{This figure compares the attribute sample sizes in the RAP-V1 dataset and the performance improvement of our method over VTB in terms of mA. The bar chart shows the sample counts for each attribute, with \textit{\textcolor{blue}{blue bars}} indicating attributes with above-average sample counts (e.g., "head") and \textit{\textcolor{orange}{orange bars}} indicating below-average sample counts (e.g., "tail"). The \textit{\textcolor{green}{green line}} represents our method's performance improvement over VTB.} 
\label{improve}
\end{figure}

\begin{figure}[!htp]
\centering
\small
\includegraphics[width=0.45\textwidth]{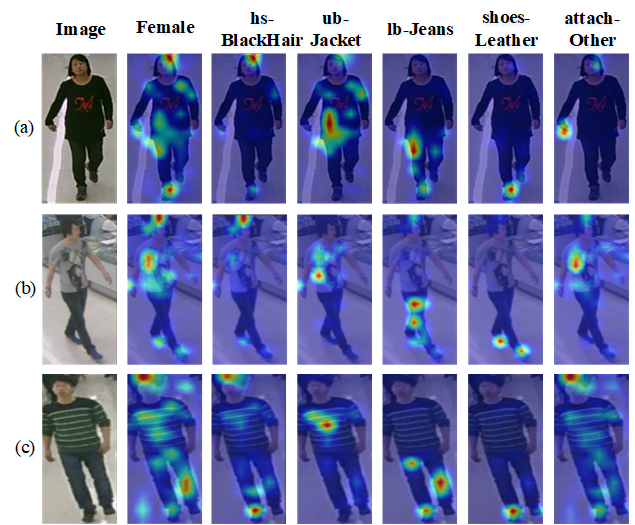}
\caption{Visualization of the response maps between pedestrian attributes and visual images. The \textcolor{red}{red} and \textcolor{blue}{blue} colors denote the high and low response.}    
\label{featuremap_vis} 
\end{figure} 

\begin{figure}[!htp]
\centering
\small
\includegraphics[width=0.45\textwidth]{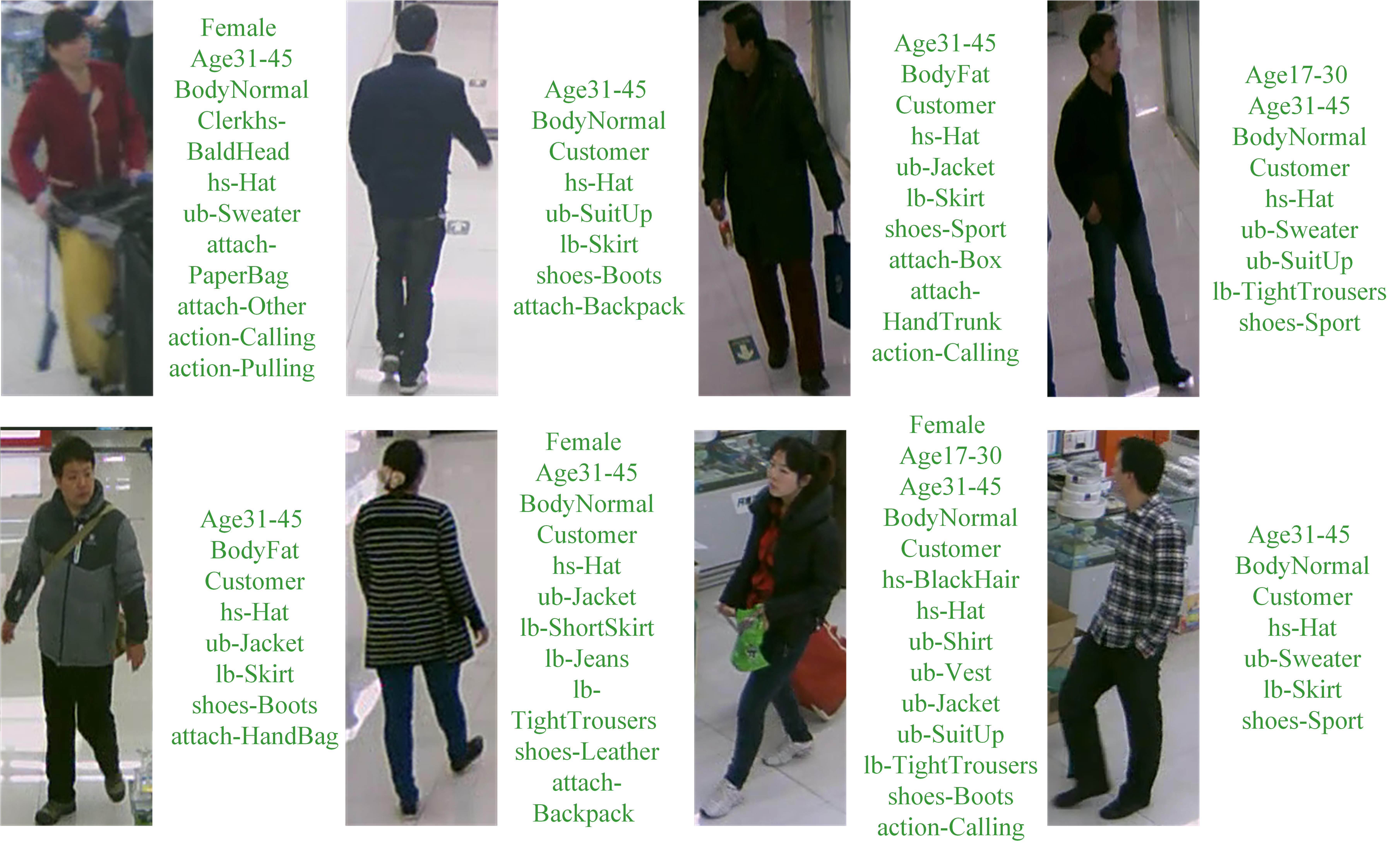}
\caption{Visualization of pedestrian attributes predicted by our proposed PromptPAR model. The \textcolor{SeaGreen4}{\emph{green}} attributes are corrected predicted ones.} 
\label{attResultsVIS} 
\end{figure}

\begin{figure}[!htp]
\centering
\small
\includegraphics[width=0.4\textwidth]{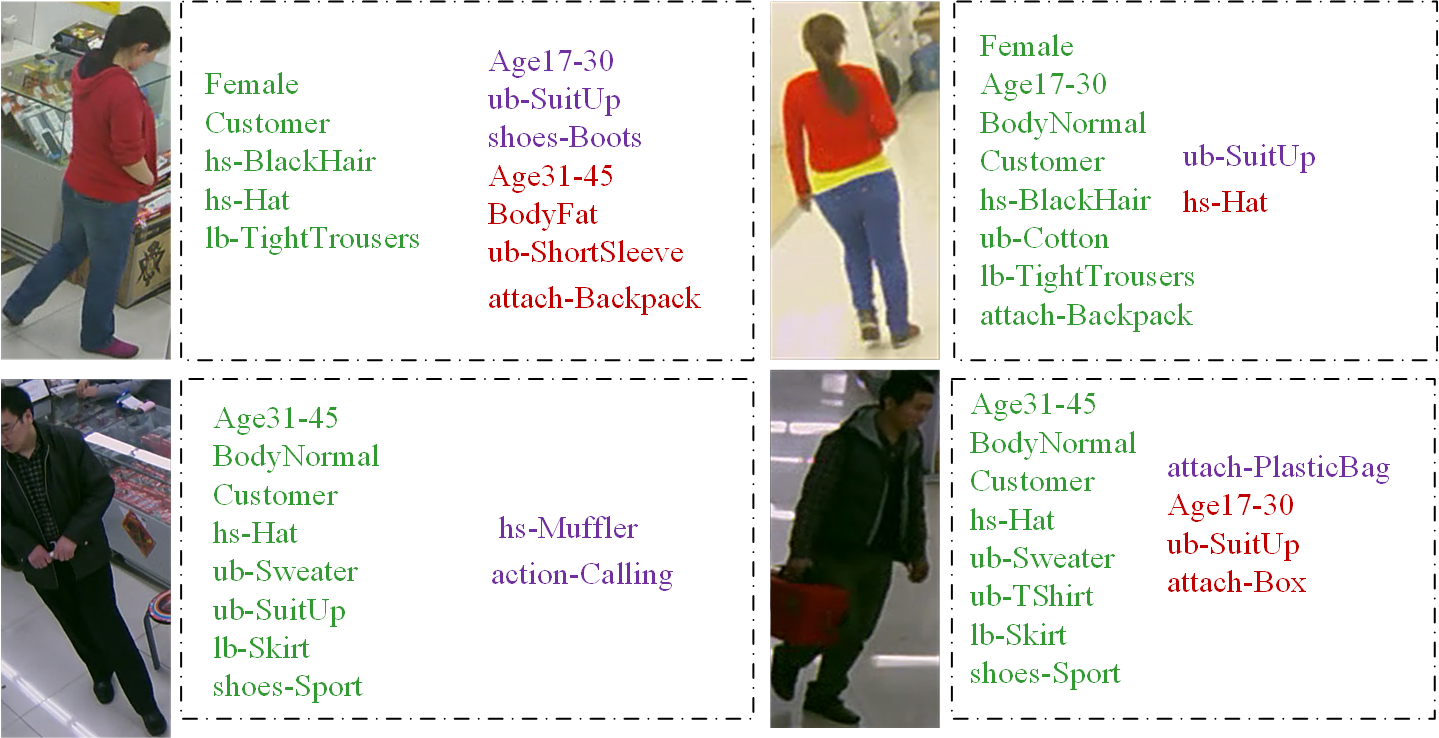}
\caption{Incorrect predictions of our proposed PromptPAR, i.e., the \textcolor{SlateBlue}{\emph{purple}} and \textcolor{DarkRed}{\emph{red}} attributes. The \textcolor{SeaGreen4}{\emph{green}} attributes are corrected predicted ones.}  
\label{label_error} 
\end{figure}

\subsection{Visualization} 
In addition to the quantitative analysis, in this sub-section, we also give a qualitative analysis to further validate the effectiveness of our proposed PromptPAR, including the visualization of feature maps and predicted human attributes. 

\textbf{Response Maps of Prompts.~} In this section, we give a visualization of the response maps between pedestrian image and prompts. Specifically, we extract attention from the last encoder of the visual encoder for visualization, and the detailed implementation can be found in our code. As shown in Fig.~\ref{respond}, We can find that previous global prompts (VPT~\cite{2022vpt}) would focus excessively on a particular region, such as the head, while ignoring detailed features in other regions, such as the body and legs, and that the parts of interest were highly similar between these prompts. After introducing our RegionPTune, it was found that our method learns better with local information while preserving global information. Based on these observations, we can conclude that our proposed RegionPTune works well in mining localized information. As a result, we achieve higher performance on several PAR benchmark datasets.

\textbf{Improvement of Imbalanced Data.~} In this section, we give a visualization of the number of attribute positive samples and the improvement in mA compared to our baseline VTB~\cite{cheng2022VTB}. As shown in Fig.~\ref{improve}, It can be found that the improvement of our method is more obvious in the tail class with fewer samples compared to the head class with more samples. This validates the effectiveness of our approach to addressing the data distribution imbalance by utilizing a large-scale pre-trained visual language model, CLIP~\cite{radford2021CLIP}, and devising a prompt tuning strategy to incorporate the extensive knowledge from CLIP~\cite{radford2021CLIP} without adjusting its parameters.

\textbf{Feature Maps.~} In this section, we give a visualization of the response maps between pedestrian image and corresponding attributes \footnote{\url{https://github.com/jacobgil/vit-explain}}. Specifically, we extract the attention from the encoder of MM-Former for visualization and the detailed implementation can be found in our code. As shown in Fig.~\ref{featuremap_vis}, we can find that our network focuses on the global view when predicting attributes like \emph{gender} (\emph{male} or \emph{female}). It can capture the local regions like the head or foot regions when predicting attributes like \emph{hair style} and \emph{shoes}. Based on these observations, we can draw the conclusion that our proposed model works well in mining the spatial relations between human attributes and images. Therefore, we achieve higher performance on multiple PAR benchmark datasets.

\textbf{Recognition Results.~}To help the readers better understand the effectiveness of our model, as shown in Fig.~\ref{attResultsVIS}, we give a visualization of predicted pedestrian attributes by our PromptPAR on the RAP-V1 dataset. More in detail, we can find that our model performs well even in \emph{motion blur, occlusion, cluttered background}, etc. These visualized results fully validated the good performance of our model in practical scenarios.

\subsection{Limitation Discussion}  
Our proposed PromptPAR achieves high performance on existing PAR benchmark datasets due to the utilization of a pre-trained big model. However, it still predicts the wrong attributes for some challenging pedestrian images, such as the attributes highlighted in \textcolor{SlateBlue}{\emph{purple}} and \textcolor{DarkRed}{\emph{red}}, as shown in Fig.~\ref{label_error}. This may be caused by the fact that our model does not consider the relations between local spatial regions of images and attributes. In addition, the running efficiency of our PromptPAR is still lower than the baseline approach. It is hard to employ such a model with a huge number of parameters in practical platforms. In our future works, we will consider treating our model as the SuperNet and distilling new lightweight neural networks for accurate and fast pedestrian attribute recognition.

\section{Conclusion and Future Work}  
In this paper, we study the pedestrian attribute recognition task from the perspective of vision-language fusion and propose a novel PromptPAR framework. Specifically, we expand the pedestrian attributes into language descriptions using the prompt engineer. The pre-trained vision-language big model CLIP is adopted to extract both the features of visual and language inputs. Then, a multi-modal Transformer is utilized to fuse the dual features effectively and a classification head is used for attribute prediction. To achieve efficient and high-performance training, we introduce the prompt tuning technique to optimize only the input prompt vectors and classification head. In other words, the weights of pre-trained big model CLIP and multi-modal fusion Transformer are all frozen. Extensive experiments on multiple PAR datasets all validated the effectiveness of our model. Our model achieves significant improvements in the zero-shot setting of PAR, which fully demonstrates the generalization of our proposed PromptPAR. 

Although good performance can be achieved using our PromptPAR model, our model can be improved from the following aspects. 
1) The knowledge distill strategy can be exploited to generate a lightweight network to achieve more efficient inference. 
2) Joint vision-language prompt tuning can be exploited for more efficient and accurate optimization. 
We leave these in the future.

\section*{Acknowledgments}
This work is supported by National Natural Science Foundation of China General Program No.62376004, the Excellent Youth Foundation of An'hui Scientific Committee under No.2208085J18, and the National Natural Science Foundation of China under Grant No.62102205.

\small{ 
\bibliographystyle{IEEEtran}
\bibliography{reference}
}
\vspace{-2.2cm}
\begin{IEEEbiography}[{\includegraphics[width=1in,height=1.25in, clip]{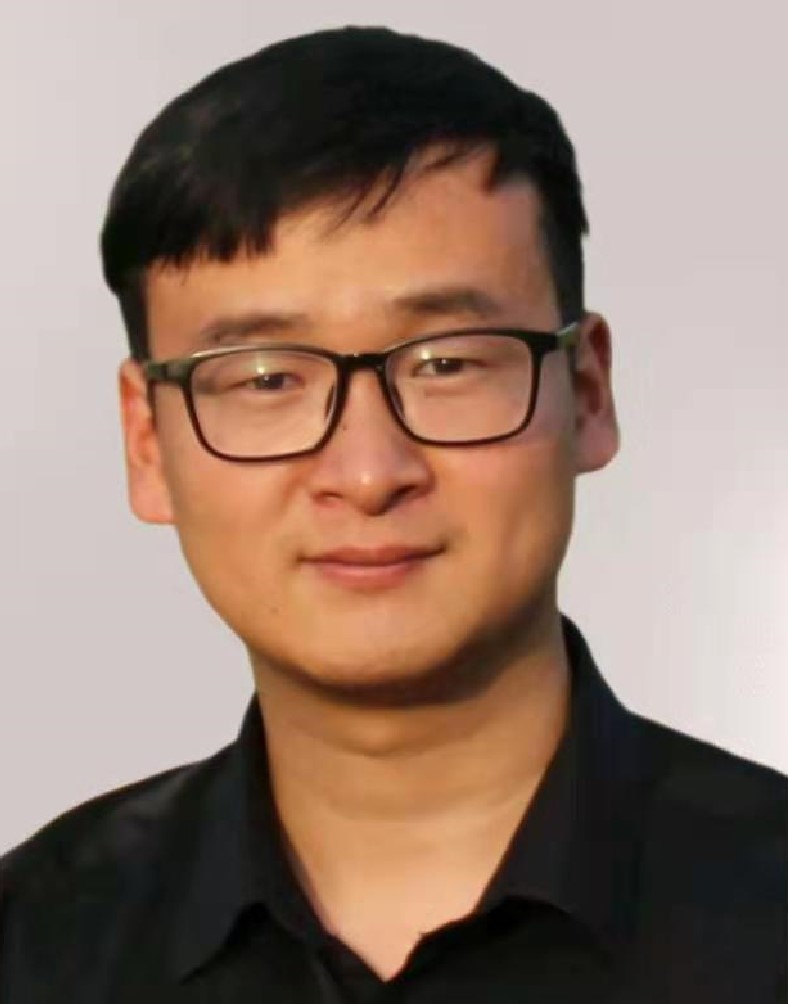}
}]{Xiao Wang} (Member, IEEE) received the B.S.  degree from West Anhui University, Lu’an, China, in 2013, and the Ph.D. degree in computer science from Anhui University, Hefei, China, in 2019. From 2015 and 2016, he was a Visiting Student with the School of Data and Computer Science, Sun Yatsen University, Guangzhou, China. He has visited the UBTECH Sydney Artificial Intelligence Centre, Faculty of Engineering, The University of Sydney, Sydney, NSW, Australia, in 2019. He is a Post-Doctoral Researcher with the Peng Cheng Laboratory, Shenzhen, China, in 2019 and 2021. He is currently an Associate Professor with the School of Computer Science and Technology, Anhui University. His current research interests include computer vision, machine learning, pattern recognition, and deep learning.
\end{IEEEbiography}
\begin{IEEEbiography}[{\includegraphics[width=1in,height=1.3in, clip]{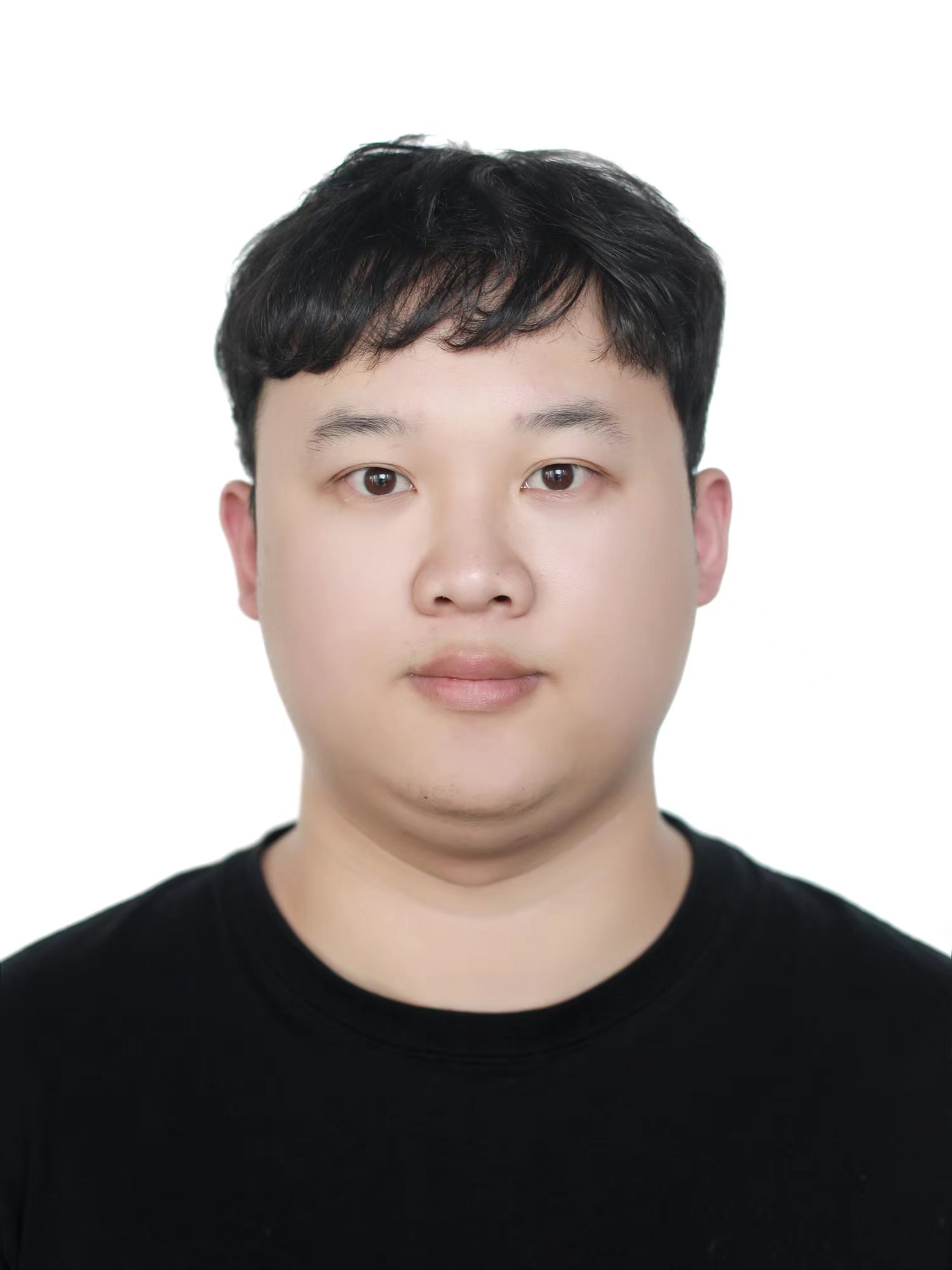}}]{Jiandong Jin} obtained a bachelor's degree from Anhui Polytechnic University, majoring in communication engineering. He is currently a graduate student at Anhui University, majoring in Artificial Intelligence. His current research interests are computer vision and pedestrian attribute recognition. 
\end{IEEEbiography}
\vspace{-1.9cm}
\begin{IEEEbiography}[{\includegraphics[width=1in,height=1.3in, clip]{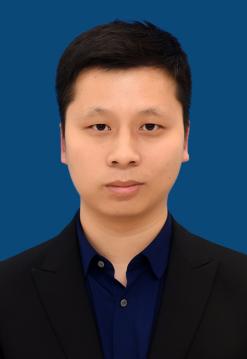}}]{Chenglong Li} received the M.S. and Ph.D. degrees from the School of Artificial Intelligence, Anhui University, Hefei, China, in 2013 and 2016, respectively. From 2014 to 2015, he worked as a Visiting Student with the School of Data and Computer Science, Sun Yat-sen University, Guangzhou, China. He was a Post-Doctoral Research Fellow with the Center for Research on Intelligent Perception and Computing (CRIPAC), National Laboratory of Pattern Recognition (NLPR), Institute of Automation, Chinese Academy of Sciences (CASIA), Beijing, China.  He is currently a Professor with the School of Artificial Intelligence, Anhui University. His research interests include computer vision and deep learning. Dr. Li was a recipient of the ACM Hefei Doctoral Dissertation Award in 2016.
\end{IEEEbiography}
\vspace{-1.9cm}
\begin{IEEEbiography}[{\includegraphics[width=1in,height=1.3in, clip]{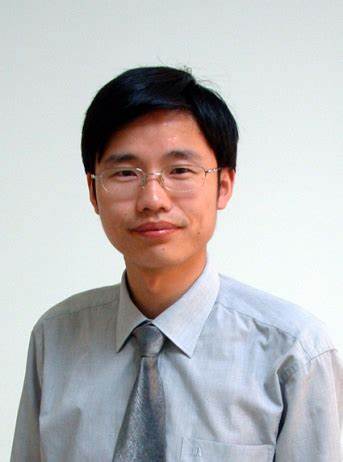}}]{Jin Tang} received the B.Eng. degree in automation and the Ph.D. degree in computer science from Anhui University, Hefei, China, in 1999 and 2007, respectively. He is currently a Professor with the School of Computer Science and Technology, Anhui University. His current research interests include computer vision, pattern recognition, machine learning, and deep learning.
\end{IEEEbiography}
\vspace{-1.9cm}
\begin{IEEEbiography}[{\includegraphics[width=1in,height=1.3in, clip]{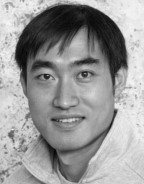}}]{Cheng Zhang} received the MSc and PhD degrees from the University of Durham in 2006 and 2011. respectively. His research interests include the study of how software design patterns are used, empirical software engineering, and the use of evidence-based software engineering techniques.
\end{IEEEbiography}
\vspace{-1.9cm}
\begin{IEEEbiography}[{\includegraphics[width=1in,height=1.3in, clip]{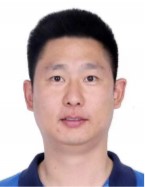}}]{Wei Wang,} Master's degree holder, Associate Senior Title, focuses on video big data research and video technology application innovation, Video Investigation Unit, Hefei Public Security Bureau.
\end{IEEEbiography}

\end{document}